\documentclass[letterpaper]{article}
\usepackage[preprint]{aaai2027}
\usepackage[hyphens]{url}
\usepackage{graphicx}
\usepackage{natbib}
\usepackage{caption}
\usepackage{booktabs}
\usepackage{amsmath,amssymb}
\usepackage{array}
\usepackage{placeins}

\title{DRIFT: Direct--Recursive Intervention-Conditioned Forecasting of ICU Physiological Trajectories}
\author{
Weixin Liu\textsuperscript{1},
Juming Xiong\textsuperscript{1},
Congning Ni\textsuperscript{2},
Yanfan Zhu\textsuperscript{1},\\
Xingtao Lin\textsuperscript{1},
Bradley A. Malin\textsuperscript{1,2},
Zhijun Yin\textsuperscript{1,2}
}
\affiliations{
\textsuperscript{1}Vanderbilt University, Nashville, Tennessee, USA\\
\textsuperscript{2}Vanderbilt University Medical Center, Nashville, Tennessee, USA
}

\newcommand{\method}{DRIFT}
\newcommand{\tftlarge}{TFT-Large}
\newcommand{\tftobj}{TFT-Large-FL}

\newcommand{\pp}{\mathrm{pp}}
\newcommand{\na}{\textemdash}
\newcommand{\zdir}{\mathbf{z}^{\mathrm{dir}}}
\newcommand{\zrec}{\mathbf{z}^{\mathrm{rec}}}
\newcommand{\zfinal}{\mathbf{z}^{\mathrm{final}}}
\newcommand{\best}[1]{\mathbf{#1}}
\newcommand{\second}[1]{\underline{#1}}

\begin{document}
\maketitle

\begin{abstract}
Many time-series forecasts depend not only on prior observations but also on actions specified during the forecast period. In intensive care units (ICUs), future vital signs and laboratory values are influenced by treatments such as vasopressors.
However, models that predict the full future sequence all at once make little use of these treatments, whereas autoregressive models can accumulate errors. We introduce \method{}, a hybrid framework in which a direct model produces the primary forecast and a recursive, action-conditioned model contributes constrained corrections. We evaluate \method{} on 6,046 admissions from MIMIC-IV and 8,345 admissions from eICU-CRD. Averaged across the 8-, 24-, and 48-hour forecast endpoints, \method{} reduces mean absolute error for mean arterial pressure (MAP) by $0.673\%$ relative to an action-conditioned Temporal Fusion Transformer (TFT-action) on MIMIC-IV and achieves the lowest corresponding error among the compared models on eICU-CRD. Although the overall accuracy improvement is modest, a MIMIC-IV audit restricted to windows in which the supplied treatment sequence was altered showed that \method{} achieved lower observed-target MAP error than TFT-action at 8 and 24 hours. Treatment-sequence alteration increased \method{}'s MAP error by $0.21$--$0.26$ mmHg more than it increased TFT-action's error, with prediction changes occurring primarily after the supplied paths diverged. In a separate robustness experiment, the MAP advantage persisted under three shared checkpoint-selection rules emphasizing overall endpoint error, MAP error, or both equally.
\end{abstract}

\section{Introduction}

Many time-series forecasting problems depend on both past observations and actions specified during the forecast period. In an intensive care unit (ICU), reliable forecasts should therefore reflect not only a patient's recent clinical history but also the treatments delivered during the forecast period. Vasopressors are especially relevant because clinicians repeatedly adjust them to maintain adequate blood pressure in critically ill patients. This setting gives rise to an action-conditioned multivariate forecasting problem involving 27 physiological measurements over 8-, 24-, and 48-hour horizons. Among these outcomes, mean arterial pressure (MAP) is particularly important because of its close relationship to vasopressor therapy and organ perfusion \citep{evans2021surviving}.

Relevant approaches include direct, recursive, and hybrid multi-step forecasting, as well as clinical world models. Direct multi-output approaches generate the full forecast horizon in a single forward pass without feeding predicted physiological values back into subsequent steps. Recursive approaches instead propagate predicted outputs or latent states over time, explicitly modeling temporal dependence but risking error accumulation during rollout. Hybrid methods seek to balance these tradeoffs \citep{green2025stratify}. For instance, ProNet divides the forecast horizon into learned segments and combines autoregressive with non-autoregressive prediction \citep{lin2023pronet}, whereas Stratify pairs base and residual forecasters drawn from recursive, direct, and direct--recursive multi-output strategies \citep{green2025stratify}. In the clinical world-model formulations cited here, the learned transition is used to roll patient states forward under candidate interventions, often for simulation or treatment planning \citep{mu2026ehrworld,wang2026chronomedicalworld}, including recent ICU work on sepsis treatment recommendation \citep{wu2026agentifying}. \method{} uses the transition differently: the direct forecaster produces the full multivariate trajectory, while the recursive transition is auxiliary, is never decoded into an independent trajectory, and can modify the direct representation only through a bounded correction.

\begin{figure*}[t]
    \centering
    \includegraphics[width=0.95\textwidth]{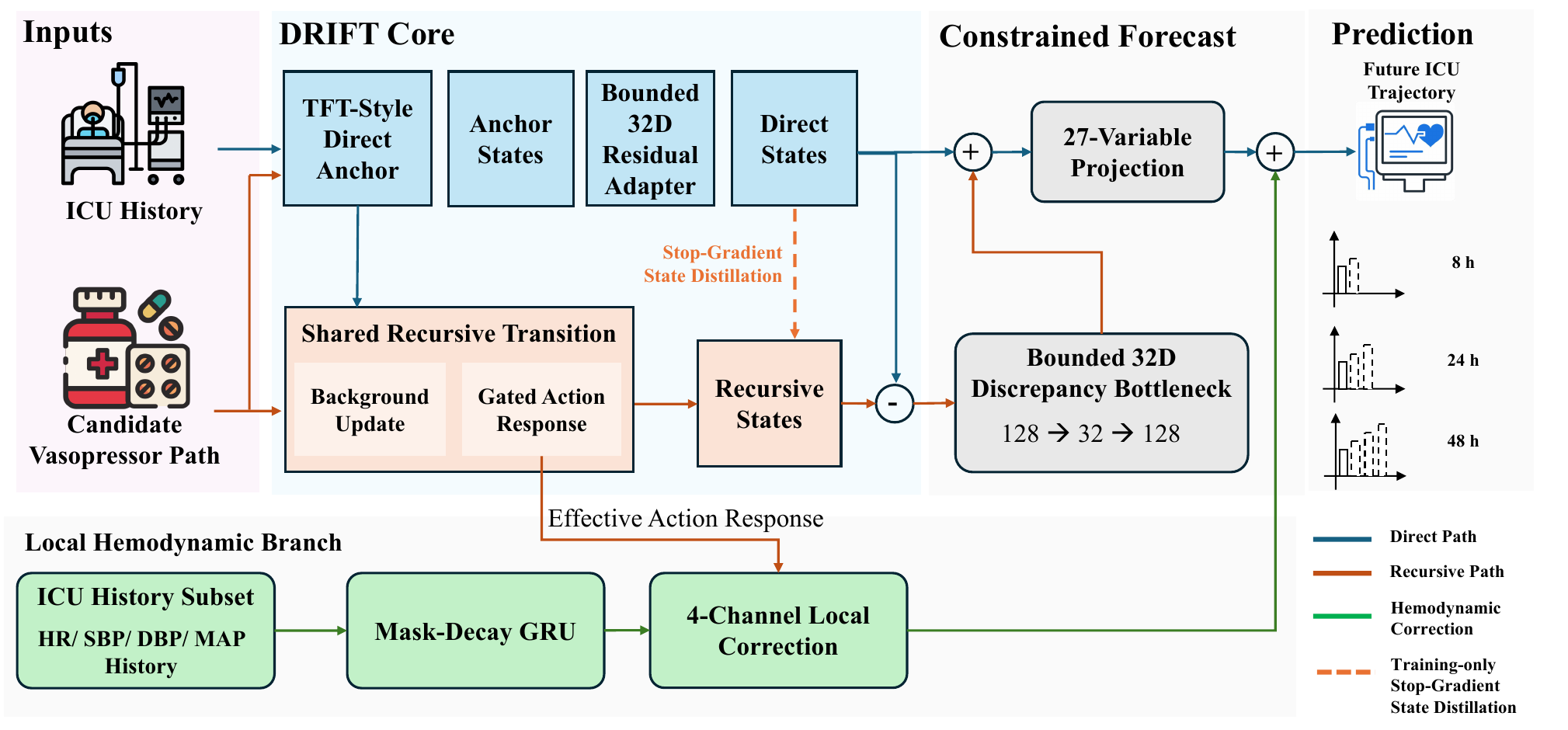}
    \caption{Overview of \method{}. A Temporal Fusion Transformer (TFT) anchor directly represents all future hours, while an action-conditioned recursive transition advances a second state one hour at a time. Stop-gradient distillation aligns the two paths, and only their bounded low-dimensional discrepancy corrects the direct state. A shared decoder predicts 27 variables, followed by a local mask-decay gated recurrent unit (GRU) correction for heart rate (HR), systolic blood pressure (SBP), diastolic blood pressure (DBP), and MAP.}
    \label{fig:drift-architecture}
\end{figure*}

We propose \method{} (Figure~\ref{fig:drift-architecture}), in which the recursive transition is an auxiliary latent correction pathway rather than an independently decoded forecasting model. A Temporal Fusion Transformer (TFT) anchor maps the patient history, supplied hourly vasopressor sequence, and known context to a direct state for each future hour \citep{lim2021tft}, while a shared transition model recursively advances a second state under the same action sequence. Stop-gradient distillation aligns the recursive states with their direct counterparts during training, and their remaining discrepancy is compressed through a bounded low-dimensional bottleneck before being added as a correction to the direct state. Thus, unlike hybrid methods that decode or combine separate direct and recursive forecasts, \method{} preserves a single shared decoder for all 27 variables and restricts the recursive path to making small, action-sensitive refinements. A local mask-decay GRU then uses observation masks and elapsed times to refine heart rate and three blood-pressure variables.

We develop \method{} on MIMIC-IV and evaluate it on eICU-CRD using independent preprocessing and training under a protocol finalized before eICU-CRD test evaluation. Across both databases, \method{} achieves a modest improvement in MAP forecasting and exhibits stronger dependence on the supplied action sequence than direct TFT baselines. Additional MIMIC-IV audits show that, when sequence substitution alters the future action path, \method{} is more accurate under the recorded sequence at 8 and 24 hours, degrades more under the substituted sequence, and changes its predictions primarily after the two paths diverge. These findings remain consistent across three shared checkpoint-selection criteria. In summary, our contributions are a direct--recursive architecture with a single decoder, cross-database evaluation of recorded-path forecasting and action-path dependence, and fine-grained audits of changed paths, temporal alignment, and checkpoint-selection robustness.

\section{Related Work}

\paragraph{Clinical time-series forecasting.}
Clinical forecasting methods must handle irregular observation and long-range temporal dependence. Missingness-aware recurrent and continuous-time models incorporate observation masks, elapsed times, or continuous dynamics to represent sparsely measured trajectories \citep{che2018grud,rubanova2019latent,kidger2020neural,shukla2021mtan}. Multi-horizon Transformer forecasters instead produce several future leads jointly and capture longer-range dependencies through attention or patch- and variable-wise representations \citep{lim2021tft,nie2023patchtst,liu2024itransformer}. These methods are evaluated mainly by how accurately they predict recorded trajectories. Our setting additionally provides a future clinical action sequence and asks whether the forecast changes when that input changes.

\paragraph{Direct, recursive, and clinical world models.}
Direct multi-output strategies do not feed predicted targets into later horizons, whereas recursive strategies reuse predicted outputs or latent states and can propagate errors. ProNet is a hybrid neural forecaster that mixes autoregressive and non-autoregressive prediction across learned horizon segments, while Stratify pairs base and residual forecasters selected from recursive, direct, and direct--recursive multi-output strategies \citep{lin2023pronet,green2025stratify}. Clinical world models and related trajectory-prediction frameworks learn patient-state transitions or generate longitudinal trajectories, sometimes under sequential interventions, including recent ICU work on sepsis treatment recommendation \citep{adam2026movingdocument,mu2026ehrworld,wang2026chronomedicalworld,yang2026clinjepa,wu2026agentifying}. \method{} instead treats the recursive transition as an auxiliary latent correction: it is never decoded independently and can modify the direct multivariate forecast only through a bounded bottleneck and one shared decoder.

\paragraph{Treatment-aware sequence modeling.}
Counterfactual treatment-response models estimate outcomes under alternative actions using explicit causal assumptions \citep{robins2000marginal,lim2018forecasting,bica2020estimating,yoon2018ganite,shalit2017estimating}; decision support under policy changes requires still stronger structure \citep{schulam2017reliable}. \method{} estimates neither a policy nor counterfactual outcomes: actions are used as predictive covariates, and the replacement analysis audits a frozen model's dependence on the supplied path. This distinction is important because changes in measurement, documentation, and site mix can alter predictive associations \citep{zech2018variable,kelly2019key,futoma2020myth}. Intervention-aware reconstruction evaluates whether a localized feature change propagates across correlated variables \citep{delibasoglu2026intervention}; our audit instead measures how replacing a clinical action sequence changes factual forecasting error.

\section{Method}

Figure~\ref{fig:drift-architecture} summarizes the model. A direct path produces the primary trajectory, an action-conditioned recursive path models hourly state changes, and constrained fusion permits only a bounded low-dimensional correction. A shared projection predicts 27 variables, followed by a four-variable hemodynamic correction.

\subsection{Problem Setup}

Each ICU admission (hereafter, stay) is represented as a sequence of one-hour bins. For stay $i$ and hour $t$, $\mathbf{x}_{i,t}\in\mathbb{R}^{D}$ contains the normalized values of the $D=27$ physiological variables after past-only filling. The binary vector $\mathbf{m}_{i,t}\in\{0,1\}^{D}$ records which variables were directly measured during that hour, and $\boldsymbol{\delta}_{i,t}\in\mathbb{R}^{D}_{+}$ records the number of hours since each variable was last measured. For example, if MAP is measured during hour $t$ but lactate is not, then $m^{\mathrm{MAP}}_{i,t}=1$ and $\delta^{\mathrm{MAP}}_{i,t}=0$, whereas $m^{\mathrm{lactate}}_{i,t}=0$ and $\delta^{\mathrm{lactate}}_{i,t}$ is the elapsed time since the preceding lactate measurement. Binary exposure to any of the five vasopressors during hour $t$ is denoted by $a_{i,t}\in\{0,1\}$.

At forecast origin $t$, the observed history $\mathcal H^{(L)}_{i,t}$ contains the most recent $L\leq48$ hourly bins ending at $t$. It includes physiology through hour $t$ and recorded action history only through hour $t-1$. The 48-hour maximum was chosen a priori to provide two days of historical context, match the longest reporting horizon, and keep the input length and computational cost fixed; it was not selected using test outcomes. When fewer than 48 hours of history are available, the model uses the shorter observed sequence, with left padding identified by validity indicators. The model additionally receives a supplied candidate action sequence $\widetilde a_{i,t:t+h-1}$ for the next $h$ hourly transitions, known time features $\boldsymbol\tau_{i,1:h}$, and static predictive context $\mathbf s_i$. For each lead $k$, it predicts
\begin{equation}
\hat{\mathbf{x}}_{i,t+k}
=
F_{\theta,k}\!\left(
\mathcal H^{(L)}_{i,t},
\widetilde a_{i,t:t+k-1},
\boldsymbol\tau_{i,1:k},
\mathbf s_i
\right),
\label{eq:forecast-task}
\end{equation}
for $k=1,\ldots,h$ and $h\in\{8,24,48\}$. Thus, the candidate action assigned to the interval from $t+k-1$ to $t+k$ may condition $\hat{\mathbf{x}}_{i,t+k}$, but actions assigned to later intervals may not. For example, when predicting physiology at hour $t+2$, the candidate action assigned to the interval from $t+1$ to $t+2$ may be used, whereas the action assigned to the later interval from $t+2$ to $t+3$ may not. One model call returns the complete trajectory $\hat{\mathbf{x}}_{i,t+1:t+h}$: the direct branch is causally masked over the available action prefix, and the recursive branch performs its internal hourly rollout. The tilde denotes an externally supplied action scenario, not an action secretly observed after the forecast origin. Retrospective factual evaluation supplies the recorded path; future physiology is never an input, and \method{} neither generates nor recommends actions.

The action information for each hour $\ell$ is encoded as the four-dimensional vector
\[
\mathbf{u}_{i,\ell}
=
[a_{i,\ell},o_{i,\ell},d_{i,\ell}/24,r_{i,\ell}/24],
\]
where $a$ indicates current exposure, $o$ marks the first hour of a new exposure episode, $d$ is the number of consecutive hours in the current active episode, and $r$ is the number of hours since the most recent exposure. Duration and time-since values are clipped at 72 hours before division by 24. For example, the first hour of an episode is encoded as $[1,1,1/24,0]$, the next uninterrupted hour as $[1,0,2/24,0]$, and the first hour after treatment stops as $[0,0,0,1/24]$. Onset can reactivate after an interruption, duration resets when exposure stops, and time since exposure is zero while exposure is active.

For forecast lead $k$, the candidate vector $\widetilde{\mathbf u}_{i,k}$ is constructed from the candidate action assigned to hour $t+k-1$ together with the preceding action history supplied to the model. Historical action vectors are shifted by one hour so that the vector for hour $\ell$ accompanies the physiological transition into hour $\ell+1$. Historical time features contain the normalized absolute ICU hour and a validity indicator for padded positions. Future time features contain the normalized relative forecast lead and the normalized absolute target ICU hour; these known features are denoted by $\boldsymbol{\tau}_{i,k}$. The two treatment-likelihood features in $\mathbf{s}_{i}$ are fitted on training data only and are used as predictive context rather than causal adjustment. We suppress the stay index on latent states when it is unambiguous.

\subsection{Direct Anchor and Recursive Transition}

The direct anchor is a point-forecast TFT-style network with 128 hidden units, one-layer encoder and decoder long short-term memory (LSTM) networks, four-head causal attention, grouped variable selection, and gated residual blocks \citep{vaswani2017attention,lim2021tft}. It maps history $\mathcal{H}^{(L)}_{i,t}$, the candidate action vectors, time features, and static context to anchor states $\mathbf{z}^{\mathrm{anchor}}_{1:h}$ and the last historical state $\mathbf{h}_{t}$. A zero-initialized adapter with a 32-dimensional bottleneck permits a bounded update,
\begin{align}
\boldsymbol{\delta}^{s}_{k}&=\alpha_{s}\tanh\!\left(P_{s,2}\operatorname{GELU}(P_{s,1}\operatorname{LN}(\mathbf{z}^{\mathrm{anchor}}_{k}))\right),\\
\zdir_{k}&=\mathbf{z}^{\mathrm{anchor}}_{k}+\boldsymbol{\delta}^{s}_{k}.
\end{align}
Here, $\operatorname{LN}$ and $\operatorname{GELU}$ denote layer normalization and the Gaussian error linear unit, respectively. The zero-initialized output projection makes the initial direct trajectory identical to the pretrained anchor trajectory.

The recursive path starts from $\zrec_{0}=\operatorname{LN}(\mathbf{h}_{t}+P_{\mathrm{init}}(\mathbf{h}_{t}))$, where $P_{\mathrm{init}}$ is a zero-initialized adapter with a 32-dimensional bottleneck, bounded by the same direct-state budget $\alpha_s$. At step $k$, shared networks produce
\begin{align}
\boldsymbol{\Delta}^{0}_{k}&=\alpha_{0}\tanh B_{\theta}([\zrec_{k-1},\boldsymbol{\tau}_{k}]),\\
\boldsymbol{\Delta}^{a}_{k}&=\alpha_{a}\tanh R_{\theta}([\zrec_{k-1},\widetilde{\mathbf u}_{i,k},\boldsymbol{\tau}_{k},\mathbf{s}_{i}]),\\
g^{a}_{k}&=\sigma G_{\theta}([\zrec_{k-1},\widetilde{\mathbf u}_{i,k},\boldsymbol{\tau}_{k},\mathbf{s}_{i}]).
\end{align}
$B_{\theta}$ and $R_{\theta}$ are layer-normalized GELU multilayer perceptrons (MLPs) with hidden width 256 and 128-dimensional outputs; $G_{\theta}$ uses a 128-dimensional GELU hidden layer and a scalar output. All three networks share parameters across forecast steps. Let
$q_{i,k}=\mathbb{1}[\sum_{\ell=1}^{k}\widetilde a_{i,t+\ell-1}>0]$.
Thus, $q_{i,k}$ depends only on the candidate future path, is zero for an all-zero path, and remains one after the first candidate exposure. The action feature vector distinguishes active, stopped, and reinitiated episodes. The transition is
\begin{equation}
\zrec_{k}=\operatorname{LN}\!\left(\zrec_{k-1}+\boldsymbol{\Delta}^{0}_{k}+q_{i,k}g^{a}_{k}\boldsymbol{\Delta}^{a}_{k}\right).
\end{equation}
The final weights of $G_{\theta}$ are initialized to zero. Its negative, trainable action-gate bias $b_a^{(0)}$, listed in Appendix Table~\ref{tab:stage-weights}, keeps the newly added response small at the start of training.

The recursive state follows the direct state through internal stop-gradient distillation \citep{hinton2015distilling},
\begin{equation}
\mathcal{L}_{\mathrm{distill}}=\mathbb{E}_{i,k}\!\left[\rho_i\frac{1}{d_z}\left\|\zrec_{i,k}-\operatorname{sg}(\zdir_{i,k})\right\|_{2}^{2}\right],
\end{equation}
where $d_z=128$ and $\rho_i$ is the number of observed historical feature entries divided by the product of $D$ and the number of valid historical hours, clipped to $[0.1,1]$. The stop-gradient operator $\operatorname{sg}$ prevents this term from updating the direct target.

\subsection{Constrained Fusion and Local Hemodynamics}

The recursive trajectory is never decoded independently. Instead, its discrepancy from the direct trajectory passes through a zero-initialized bottleneck:
\begin{align}
\mathbf{d}_{k}&=\zrec_{k}-\zdir_{k},\\
g^{c}_{k}&=\sigma G_c([\operatorname{LN}(\mathbf d_k),\boldsymbol{\tau}_k,\mathbf s_i]),\\
\mathbf{c}_{k}&=g^{c}_{k}\,\alpha_{c}\tanh\!\left(W_{c,2}\operatorname{GELU}(W_{c,1}\mathbf{d}_{k})\right),\\
\zfinal_{k}&=\zdir_{k}+\mathbf{c}_{k}.
\end{align}
The correction maps $128\rightarrow32\rightarrow128$. $G_c$ has a 64-dimensional GELU hidden layer and a scalar output; its final weights and the 32-to-128 projection are initialized to zero. The gate bias has the trainable initialization $b_c^{(0)}$ listed in Appendix Table~\ref{tab:stage-weights}. The bottleneck therefore limits recursive information to a low-dimensional adjustment rather than a second full decoder.

Vasopressors primarily motivate specialization of hemodynamic variables, including MAP as a common clinical target \citep{evans2021surviving}. A 64-dimensional mask-decay gated recurrent unit (GRU) processes HR, SBP, DBP, and MAP histories using their values, masks, and elapsed times \citep{cho2014gru,che2018grud}. Its final state, future time features, and static context produce a gated, future-action-independent base correction. The effective recursive action response $\mathbf{e}^{a}_{i,k}=q_{i,k}g^{a}_{k}\boldsymbol{\Delta}^{a}_{k}$ is separately projected to four channels. If $S_H^{\top}$ inserts those channels into the 27-variable output, the prediction is
\begin{equation}
\hat{\mathbf{x}}_{i,t+k}=O_{\phi}(\zfinal_{i,k})+S_H^{\top}\!\left(\mathbf{c}^{H,0}_{i,k}+\mathbf{c}^{H,a}_{i,k}\right).
\end{equation}
Both local output projections are zero initialized and bounded by $\alpha_H$. The local base gate uses the negative, trainable initialization $b_H^{(0)}$ listed in Appendix Table~\ref{tab:stage-weights}. The shared projection $O_{\phi}$ is inherited from the direct TFT; no second 27-variable decoder is added.

We denote the direct-state adapter, background transition, action response, discrepancy, and local-output bounds by $\alpha_s$, $\alpha_0$, $\alpha_a$, $\alpha_c$, and $\alpha_H$, respectively, and call them \emph{bounded residual budgets}. Because latent residuals operate on layer-normalized states and local corrections operate in normalized output space, these dimensionless architectural bounds restrict newly added pathways from overriding the direct anchor. Appendix Table~\ref{tab:residual-budget-sensitivity} reports their frozen defaults and a joint-scale sensitivity analysis.

\subsection{Training Objective and Curriculum}

The forecast loss combines feature-weighted trajectory mean squared error (MSE), terminal endpoint MSE, and raw-scale Huber loss for HR, SBP, DBP, and MAP. MAP, lactate, and creatinine receive weight two in the trajectory term, and MAP receives the largest hemodynamic weight. Additional terms preserve the contemporaneous anchor forecast, align recursive and direct states, penalize pre-exposure action responses, and limit the magnitudes of the adapter, discrepancy, and local corrections:
\begin{equation}
\begin{split}
\mathcal{L}={}&\lambda_{\mathrm{traj}}\mathcal{L}_{\mathrm{traj}}+\lambda_{\mathrm{end}}\mathcal{L}_{\mathrm{end}}+\lambda_{\mathrm{hemo}}\mathcal{L}_{\mathrm{hemo}}\\
&+\lambda_{\mathrm{anchor}}\mathcal{L}_{\mathrm{anchor}}+\lambda_{\mathrm{distill}}\mathcal{L}_{\mathrm{distill}}\\
&+\lambda_{\mathrm{pre}}\mathcal{L}_{\mathrm{pre}}+\lambda_{\mathrm{budget}}\mathcal{L}_{\mathrm{budget}}.
\end{split}
\end{equation}
Training follows a four-stage curriculum that activates the direct-state adapter and local hemodynamic base, then the recursive transition and discrepancy path, then the local action projection, and finally low-rate updates to the anchor output blocks. Zero initialization preserves the pretrained anchor at the start of adaptation, and each stage opens one bounded pathway at a time. Appendix~\ref{app:training-details} gives the objective weights and complete stage schedule. The frozen $1.0\times$ residual budgets define the primary model, and joint-scale sensitivity is reported as an additional MIMIC-IV test-set analysis.

\section{Experimental Setup}

\subsection{Optimization and Checkpoint Selection}

\method{} is trained for 20 fixed epochs with 500 minibatches per epoch and batch size 192. One horizon from $\{8,16,24,48\}$ is sampled per minibatch with probabilities $(0.30,0.15,0.30,0.25)$. AdamW uses a learning rate of $10^{-4}$ for new modules and $10^{-5}$ for the anchor blocks unfrozen in the final curriculum stage, with weight decay $10^{-4}$ and global gradient clipping at 0.5. Training uses bfloat16 (BF16) autocast with 32-bit floating-point (FP32) loss computation and no scheduler, warmup, or early stopping. Checkpoint selection uses factual validation forecasting only; neither action-replacement diagnostics nor test outcomes enter selection.

To test whether model-specific validation scores affected the comparison, we conducted a separate MIMIC-IV robustness experiment. \method{} and TFT-action were retrained for seeds 42--44 with every epoch checkpoint retained. Before test inference, one checkpoint per model and seed was frozen under three shared criteria combining within-run normalized endpoint MSE and MAP mean absolute error (MAE): endpoint-heavy ($0.80/0.20$), equal ($0.50/0.50$), and MAP-heavy ($0.20/0.80$). The same test roster, horizons, donor mappings, changed-window definitions, bootstrap, and multiplicity correction were used; the full shared-selection protocol and results appear in the appendices.

\subsection{Datasets and Protocols}

The primary benchmark uses MIMIC-IV v3.1 \citep{johnson2023mimiciv,johnson2024mimiciv31}. Fixed eligibility rules yield 40,301 stays, split by patient into 28,210 training, 6,045 validation, and 6,046 test stays. Each stay is represented by 27 variables over the first 72 ICU hours. Missing measurements are filled using only earlier values from the same stay; any remaining leading gaps use training-set medians, while observation masks and elapsed times are retained. The action is hourly exposure to any of five vasopressors. Appendices~\ref{app:cohort-flow}--\ref{app:action-sparsity} provide complete cohort and preprocessing definitions.

The eICU-CRD v2.0 replication \citep{pollard2018eicu,pollard2019eicu20} uses patient-disjoint training, validation, and test splits of 38,633, 8,115, and 8,345 stays. All test hospitals and hospital--ward pairs also occur in training, so the replication is cross-database but not site-held-out. Source-specific mappings preserve the MIMIC grid, targets, masks, time features, and action alignment; positive infusion snapshots use a fixed 120-minute carry. Preprocessing is fitted only on eICU training data, and no MIMIC-trained weights are transferred. Architectures, validation-selected checkpoints, evaluation rules, donor construction, estimands, bootstrap, and hypothesis families are fixed independently of test outcomes. Appendices~\ref{app:eicu-data-provenance}--\ref{app:eicu-freeze} provide the full replication protocol.

MIMIC-IV supports model development and component analysis, whereas eICU provides frozen-protocol cross-database replication. PatchTST and iTransformer appear only in the eICU benchmark.

\subsection{Baselines, Metrics, and Inference}

The primary study compares Persistence, GRU \citep{cho2014gru}, Transformer \citep{vaswani2017attention}, mask-decay GRU \citep{che2018grud}, an action-conditioned TFT (TFT-action), a TFT without dynamic action \citep{lim2021tft}, and two direct controls. The capacity-matched \tftlarge{} has 1,586,047 parameters versus 1,551,342 for \method{}, while the forecast-loss-transfer control \tftobj{} uses the principal forecast losses in the direct architecture without the recursive, distillation, discrepancy, local, or preservation pathways. PatchTST-action \citep{nie2023patchtst} and iTransformer-action \citep{liu2024itransformer} appear only in the frozen eICU-CRD benchmark. On MIMIC-IV, we additionally evaluate a literature-inspired direct--recursive multi-output comparator with fixed eight-hour forecast blocks (DirRecMO-8) \citep{lin2023pronet,green2025stratify}; its strengthened training protocol and complete results appear in Appendix~\ref{app:dirrecmo8}. Neural models use seeds 42, 43, and 44 with recorded-path validation selection; the seven MIMIC-IV ablations and their operator definitions appear in Appendices~\ref{app:ablation-full}--\ref{app:ablation-operators}.

The primary endpoint is unweighted normalized MSE across all 27 variables at the terminal forecast hour. We also report terminal-hour MAP MAE in mmHg. Observed-only metrics use only directly measured target entries, whereas integrated metrics average stay-level summaries across all 48 forecast hours on a fixed 48-hour-eligible cohort. Horizon tables report descriptive window-level means, whereas paired inference first summarizes windows within each stay and then weights reporting horizons equally; inferential differences therefore need not equal subtractions of displayed table means. Comparisons use identical stays and windows. We draw 2,000 hierarchical bootstrap samples over seeds and stays with plus-one two-sided probabilities \citep{efron1994bootstrap}. The eight prespecified eICU contrasts form three correction families: two standard-MAP contrasts, two observed-MAP contrasts, and four action-gap contrasts \citep{benjamini1995controlling}. Intervals condition on trained checkpoints and shared cohorts rather than treating seeds, horizons, or donors as independent clinical replications.

\subsection{Action-Path Replacement}

Five fixed one-to-one, no-self donor mappings are constructed within exact time-mask and transition-mask strata, without using physiology, outcomes, predictions, diagnoses, severity, or model error. Full-stream replacement substitutes historical action context and the future path. Future-only replacement preserves recipient history, substitutes future raw exposure, and rebuilds onset, duration, and time-since features. The action gap is shifted-path error minus recorded-path error; a positive value indicates that the model depends on the supplied path. Because many sparse paths remain all-zero after replacement, we report path-change prevalence beside each gap. Larger gaps do not imply accurate alternative-treatment simulation or better forecasting.

Full-stream replacement tests the complete encoded action context, whereas future-only replacement isolates the candidate segment with recipient history fixed. The observed outcome never changes and may be incompatible with the donor action; the audit measures input dependence, not shifted-trajectory validity.

After completing the primary analyses, we conducted an additional MIMIC-IV audit using frozen \method{} and TFT-action checkpoints and the same five donor mappings. A \emph{changed window} is one in which future-only substitution changes at least one hourly exposure. On this subset, we compare recorded-path observed-target MAP MAE and shifted-minus-recorded gaps, and align absolute prediction changes to the first action-path divergence. The audit uses 5,000 paired hierarchical bootstrap samples over stays and seeds with Benjamini--Hochberg correction within each MAP family. No checkpoint was trained or selected for these changed-window or temporal analyses, and eICU was not accessed.

\section{Results}

\subsection{Forecasting Across Databases}

\begin{table*}[t]
\centering
{\small
\setlength{\tabcolsep}{1mm}
\begin{tabular}{@{}lcccccc@{}}
\toprule
& \multicolumn{2}{c}{8 h} & \multicolumn{2}{c}{24 h} & \multicolumn{2}{c}{48 h}\\
\cmidrule(lr){2-3}\cmidrule(lr){4-5}\cmidrule(lr){6-7}
Model & E-MSE & MAP & E-MSE & MAP & E-MSE & MAP\\
\midrule
\multicolumn{7}{@{}l}{\textbf{MIMIC-IV}}\\
\midrule
Persistence
& 1.9167
& 11.024
& 1.8859
& 12.400
& 1.1287
& 13.643\\
GRU-action
& $1.6280\!\pm\!0.0353$
& $9.884\!\pm\!0.147$
& $1.4362\!\pm\!0.1627$
& $11.856\!\pm\!0.180$
& $1.5195\!\pm\!0.2139$
& $12.633\!\pm\!0.312$\\
Transformer-action
& $1.0750\!\pm\!0.0010$
& $9.034\!\pm\!0.011$
& $0.6107\!\pm\!0.0027$
& $10.008\!\pm\!0.011$
& $0.7812\!\pm\!0.0062$
& $10.899\!\pm\!0.021$\\
TFT-action
& $1.0635\!\pm\!0.0023$
& $8.995\!\pm\!0.027$
& $\second{0.5937\!\pm\!0.0009}$
& $\second{9.966\!\pm\!0.023}$
& $\second{0.7114\!\pm\!0.0008}$
& $10.660\!\pm\!0.020$\\
\tftlarge{}
& $\second{1.0622\!\pm\!0.0003}$
& $8.997\!\pm\!0.014$
& $0.5940\!\pm\!0.0008$
& $9.975\!\pm\!0.014$
& $0.7131\!\pm\!0.0019$
& $\second{10.652\!\pm\!0.029}$\\
\tftobj{}
& $1.0920\!\pm\!0.0021$
& $\second{8.942\!\pm\!0.016}$
& $0.6112\!\pm\!0.0019$
& $10.127\!\pm\!0.062$
& $0.7298\!\pm\!0.0028$
& $10.796\!\pm\!0.055$\\
\textbf{\method{} (ours)}
& $\best{1.0620\!\pm\!0.0008}$
& $\best{8.883\!\pm\!0.005}$
& $\best{0.5924\!\pm\!0.0016}$
& $\best{9.894\!\pm\!0.005}$
& $\best{0.7106\!\pm\!0.0016}$
& $\best{10.613\!\pm\!0.015}$\\
\midrule
\multicolumn{7}{@{}l}{\textbf{eICU-CRD}}\\
\midrule
Persistence
& 0.4246
& 10.867
& 0.7782
& 12.701
& 1.1344
& 14.182\\
GRU-action
& $\second{0.3468\!\pm\!0.0013}$
& $9.097\!\pm\!0.010$
& $0.5983\!\pm\!0.0012$
& $10.502\!\pm\!0.025$
& $0.7976\!\pm\!0.0005$
& $11.398\!\pm\!0.029$\\
Transformer-action
& $0.3488\!\pm\!0.0032$
& $9.077\!\pm\!0.010$
& $\second{0.5951\!\pm\!0.0009}$
& $\second{10.426\!\pm\!0.006}$
& $\second{0.7931\!\pm\!0.0016}$
& $11.340\!\pm\!0.009$\\
TFT-action
& $0.3479\!\pm\!0.0012$
& $9.087\!\pm\!0.009$
& $0.5972\!\pm\!0.0012$
& $10.448\!\pm\!0.011$
& $0.7949\!\pm\!0.0007$
& $\second{11.336\!\pm\!0.018}$\\
\tftlarge{}
& $0.3470\!\pm\!0.0010$
& $9.085\!\pm\!0.016$
& $0.5968\!\pm\!0.0013$
& $10.442\!\pm\!0.023$
& $0.7977\!\pm\!0.0031$
& $11.347\!\pm\!0.026$\\
\tftobj{}
& $0.3727\!\pm\!0.0015$
& $\second{9.038\!\pm\!0.008}$
& $0.6108\!\pm\!0.0006$
& $10.469\!\pm\!0.010$
& $0.8086\!\pm\!0.0008$
& $11.360\!\pm\!0.017$\\
\textbf{\method{} (ours)}
& $\best{0.3466\!\pm\!0.0008}$
& $\best{9.035\!\pm\!0.004}$
& $\best{0.5947\!\pm\!0.0012}$
& $\best{10.412\!\pm\!0.013}$
& $\best{0.7927\!\pm\!0.0008}$
& $\best{11.302\!\pm\!0.019}$\\
\bottomrule
\end{tabular}
}
\caption{Descriptive terminal-hour forecasting on all valid evaluation windows for models evaluated in both databases, reported as window-level mean $\pm$ seed standard deviation within each horizon, where h denotes hours. E-MSE denotes normalized endpoint mean squared error across all 27 physiological variables, and MAP denotes mean absolute error for mean arterial pressure in mmHg. E-MSE is normalized using each database's training statistics and is therefore comparable only within a database. Boldface and underlining mark the best and second-best displayed values, respectively; complete rankings appear in Appendices~\ref{app:primary-benchmark} and~\ref{app:eicu-ranking}. Inferential claims use stay-paired comparisons rather than table rank alone.}
\label{tab:cross-dataset-horizon}
\end{table*}

Table~\ref{tab:cross-dataset-horizon} reports the common model set. On MIMIC-IV, \method{} reduces endpoint MSE by $0.154\%$ relative to TFT-action (difference $-0.001215$, 95\% confidence interval (CI) $[-0.002718,-0.000212]$) and MAP MAE by $0.673\%$ ($-0.0667$ mmHg, CI $[-0.0828,-0.0478]$); both have two-sided bootstrap probabilities $p_{\mathrm{boot}}\leq0.001$. MAP is also lower than \tftlarge{} by $0.0634$ mmHg, whereas the endpoint-MSE interval includes zero.

On eICU, \method{} has the lowest horizon-averaged MAP MAE (10.249 mmHg) among 12 neural architectures and Persistence, improving on TFT-action and \tftlarge{} by 0.0412 and 0.0422 mmHg. A secondary comparison with the numerical runner-up, Transformer-action, gives $-0.0314$ mmHg (95\% CI $[-0.0466,-0.0157]$). All prespecified standard and observed MAP contrasts favor \method{} after within-family correction. Endpoint MSE was not prespecified for inference, and several multivariate trajectory intervals include zero; the cross-database evidence therefore supports a narrow MAP advantage, not universal improvement across all 27 variables.

\subsection{Observed Targets, Long Trajectories, and Additional Audits}

\begin{table}[t]
\centering
{\small
\setlength{\tabcolsep}{0.65mm}
\begin{tabular}{@{}lccc@{}}
\toprule
Horizon & Rec. & Gap & Resp. \\
\midrule
8 h
& \shortstack{$-0.096$\\$[-0.125,-0.068]$}
& \shortstack{$+0.254$\\$[+0.222,+0.285]$}
& \shortstack{$+0.574$\\$[+0.521,+0.621]$} \\
24 h
& \shortstack{$-0.061$\\$[-0.097,-0.026]$}
& \shortstack{$+0.264$\\$[+0.174,+0.337]$}
& \shortstack{$+0.802$\\$[+0.622,+0.980]$} \\
48 h
& \shortstack{$-0.038$\\$[-0.076,+0.002]$}
& \shortstack{$+0.211$\\$[+0.123,+0.302]$}
& \shortstack{$+0.771$\\$[+0.508,+0.970]$} \\
\bottomrule
\end{tabular}
}
\caption{Additional post-primary MIMIC-IV changed-window audit. Entries are \method{} minus TFT-action MAP differences in mmHg with 95\% paired bootstrap intervals. Rec. denotes recorded-path observed-target error, Gap denotes the shifted-minus-recorded error gap, and Resp. denotes absolute prediction response at and after path divergence. Negative Rec. values favor \method{}; positive Gap and Resp. values indicate larger values for \method{}.}
\label{tab:changed-path-main}
\end{table}

\begin{table*}[t]
\centering
{\small
\setlength{\tabcolsep}{1.2mm}
\begin{tabular}{@{}lrrrr@{}}
\toprule
Removed component
& E-MSE $\uparrow$
& MAP $\uparrow$
& E-gap $\downarrow$
& MAP-gap $\downarrow$ \\
\midrule
Recursive path
& $+0.022\%$
& $+0.129\%$
& $+0.054\pp$
& $+0.493\pp$ \\
State distillation
& $+0.013\%$
& $+0.148\%$
& $+0.002\pp$
& $-0.013\pp$ \\
Recursive action timing
& $+0.001\%$
& $+0.001\%$
& $+0.005\pp$
& $+0.022\pp$ \\
Recursive response
& $+0.007\%$
& $+0.028\%$
& $+0.036\pp$
& $+0.312\pp$ \\
Hemodynamic pathway
& $+0.012\%$
& $+0.182\%$
& $+0.011\pp$
& $-0.126\pp$ \\
State adapter
& $+0.008\%$
& $+0.005\%$
& $+0.006\pp$
& $+0.033\pp$ \\
Bottleneck discrepancy correction
& $+0.020\%$
& $+0.064\%$
& $+0.020\pp$
& $+0.019\pp$ \\
\bottomrule
\end{tabular}
}
\caption{Retrained MIMIC-IV ablations averaged over the 8-, 24-, and 48-hour horizons. E-MSE denotes normalized endpoint mean squared error across all 27 physiological variables; MAP denotes mean absolute error for mean arterial pressure; E-gap and MAP-gap denote the full-stream shifted-minus-recorded action gaps for endpoint MSE and MAP mean absolute error, respectively; pp denotes percentage points. The upward arrows identify columns reporting error increases after component removal, whereas the downward arrows identify columns reporting decreases in action gap after removal. Positive error increases indicate worse forecasting, and positive gap decreases indicate weaker action-path dependence. Operator definitions appear in Appendix~\ref{app:ablation-operators}.}
\label{tab:ablation-main}
\end{table*}

Secondary paired analyses, reported fully in the appendices, show that the MAP advantage over TFT-action persists under observed-only and integrated 1--48-hour evaluation in both databases. The MIMIC-IV full-stream and future-only difference-in-gap values are $+0.0264$ and $+0.0255$ mmHg, and the corresponding eICU values are $+0.0166$ and $+0.0152$ mmHg. Because only 16.6--22.4\% of complete streams change, no-op assignments attenuate these averages. The less accurate \tftobj{} has still larger gaps, confirming that sensitivity alone is not forecast quality. Multivariate improvements are not uniform across all 27 variables.

Table~\ref{tab:changed-path-main} summarizes the additional changed-window audit. \method{} lowers recorded-path observed MAP MAE versus TFT-action by 0.096 and 0.061 mmHg at 8 and 24 hours; the 48-hour interval includes zero. Its difference-in-gap remains positive at all horizons ($+0.254$, $+0.264$, and $+0.211$ mmHg). Before the first action divergence, prediction changes are at most 1.1\% of at/post-divergence changes for either model; afterward, \method{}'s MAP response is 24--31\% larger. Thus, \method{} is more accurate under the recorded path at short and intermediate horizons and responds more strongly after the supplied path changes.

\paragraph{Robustness to checkpoint selection.}
Under endpoint-heavy, equal, and MAP-heavy checkpoint selection, \method{} retained lower MAP MAE than TFT-action at every horizon (8 hours: $0.087$--$0.104$ mmHg; 24 hours: $0.047$--$0.054$; 48 hours: $0.037$--$0.042$), with all intervals excluding zero. Changed-window factual and difference-in-gap estimates also favored \method{} throughout, although multivariate MSE was not uniformly significant at 8 hours. Against the strengthened action-conditioned DirRecMO-8 comparator, \method{} reduced MAP MAE by $0.065$--$0.098$, $0.083$--$0.251$, and $0.084$--$0.305$ mmHg at 8, 24, and 48 hours across the three criteria; all nine MAP intervals and all nine terminal-MSE intervals excluded zero (Benjamini--Hochberg-adjusted $q=0.001161$; Appendix~\ref{app:dirrecmo8}). Complete shared-selection results appear in Appendix~\ref{app:checkpoint-robustness}, and additional pathway diagnostics appear in Appendices~\ref{app:trajectory-details}--\ref{app:utilization}.

\subsection{Ablations and Residual-Budget Sensitivity}

Table~\ref{tab:ablation-main} separates contributions to forecasting accuracy from contributions to action-path dependence. Removing any component increases both endpoint MSE and MAP MAE. The hemodynamic pathway and state distillation produce the largest MAP-accuracy losses when removed, whereas removing the complete recursive path or its explicit action response causes the largest reductions in MAP action gap. The smaller effects of action-timing features and the state adapter indicate that the principal action dependence arises from the recursive response and its constrained integration with the direct forecast.

Across selected forecast leads, the final--direct MSE remained below $0.0014$, while the correction RMS decreased from $0.0188$ at lead 1 to $0.0076$ at lead 48. This confirms that the recursive pathway remains a small adjustment to the direct forecast rather than becoming a competing predictor (Appendix~\ref{app:trajectory-details}).

\begin{table}[t]
\centering
{\small
\setlength{\tabcolsep}{1mm}
\begin{tabular}{@{}ccc@{}}
\toprule
Scale & MAP $\Delta$ & Gap $\Delta$ \\
\midrule
$0.5\times$
& \shortstack{$-0.0639$\\$[-0.0793,-0.0474]$}
& \shortstack{$+0.0296$\\$[+0.0226,+0.0353]$} \\
$1.0\times$
& \shortstack{$-0.0667$\\$[-0.0828,-0.0478]$}
& \shortstack{$+0.0255$\\$[+0.0163,+0.0387]$} \\
$2.0\times$
& \shortstack{$-0.0703$\\$[-0.0870,-0.0531]$}
& \shortstack{$+0.0348$\\$[+0.0283,+0.0434]$} \\
\bottomrule
\end{tabular}
}
\caption{Residual-budget sensitivity on MIMIC-IV, averaged over the 8-, 24-, and 48-hour horizons. Entries are differences from TFT-action in mmHg with 95\% paired intervals. MAP $\Delta$ denotes terminal MAP-error difference; Gap $\Delta$ denotes the future-only shifted-minus-recorded error-gap difference. Negative MAP $\Delta$ and positive Gap $\Delta$ favor \method{}.}
\label{tab:budget-sensitivity-main}
\end{table}

Table~\ref{tab:budget-sensitivity-main} shows that all three scales retained lower terminal MAP error and stronger future-only action-path dependence. These variants were not used for configuration or checkpoint selection, and the frozen $1.0\times$ model remains primary; complete results appear in Appendix Table~\ref{tab:residual-budget-sensitivity}.

\paragraph{Ethics and data governance.}
This retrospective secondary analysis used de-identified MIMIC-IV and eICU-CRD data under their data-use agreements; no restricted patient data were provided to external services or generative AI tools.

\section{Conclusion}

\method{} combines a direct multistep forecaster with an action-conditioned recursive transition that can modify the direct representation only through bounded corrections. Across MIMIC-IV and an independently trained eICU evaluation, it provides a small, consistent improvement in MAP forecasting. In windows where the supplied future action path changes, \method{} is more accurate under the recorded path at 8 and 24 hours and changes its predictions primarily after path divergence. The MAP advantage persists under shared checkpoint-selection rules and against the strengthened DirRecMO-8 comparator; ablations identify distinct contributions to forecasting accuracy and action-path dependence, while residual-budget scaling preserves both conclusions.

These retrospective results do not establish causal treatment effects or clinical benefit. Binary vasopressor exposure omits dose and co-interventions, and the eICU evaluation is not site-held-out. Future work should evaluate richer action representations, predictive uncertainty, and prospective or site-held-out cohorts.

\FloatBarrier
\bibliography{references}

\clearpage
\appendix
\makeatletter
\@addtoreset{table}{section}
\@addtoreset{figure}{section}
\@addtoreset{equation}{section}
\makeatother
\renewcommand{\thetable}{\thesection.\arabic{table}}
\renewcommand{\thefigure}{\thesection.\arabic{figure}}
\renewcommand{\theequation}{\thesection.\arabic{equation}}

These appendices provide complete benchmarks, cross-database paired analyses, cohort and preprocessing details, action-path audits including additional changed-window accuracy and first-divergence temporal alignment, unified checkpoint-selection robustness, a strengthened action-conditioned DirRecMO-8 comparison, training and ablation specifications, additional analyses, implementation checks, and the eICU Collaborative Research Database (eICU-CRD) replication under the frozen evaluation protocol. Terminology and identifiers follow the main text. MIMIC-IV denotes Medical Information Mart for Intensive Care IV. ICU, MSE, MAE, MAP, SD, and CI denote intensive care unit, mean squared error, mean absolute error, mean arterial pressure, standard deviation, and confidence interval, respectively. HR, SBP, and DBP denote heart rate, systolic blood pressure, and diastolic blood pressure. GRU and TFT denote gated recurrent unit and Temporal Fusion Transformer, and TFT-Large-FL is the forecast-loss-transfer control.

\section{Complete Benchmarks and Reporting Conventions}
\label{app:benchmarks}

\subsection{Complete Primary-Cohort Benchmark}
\label{app:primary-benchmark}

The following table preserves the full primary-cohort ranking condensed in the main cross-database table.

\begin{table*}[!t]
\centering
\scriptsize
\setlength{\tabcolsep}{3.1pt}
\begin{tabular}{lcccccc}
\toprule
& \multicolumn{2}{c}{8 hours} & \multicolumn{2}{c}{24 hours} & \multicolumn{2}{c}{48 hours}\\
\cmidrule(lr){2-3}\cmidrule(lr){4-5}\cmidrule(lr){6-7}
Model & Endpoint MSE $\downarrow$ & MAP MAE (mmHg) $\downarrow$ & Endpoint MSE $\downarrow$ & MAP MAE (mmHg) $\downarrow$ & Endpoint MSE $\downarrow$ & MAP MAE (mmHg) $\downarrow$\\
\midrule
Persistence & 1.9167 & 11.024 & 1.8859 & 12.400 & 1.1287 & 13.643\\
GRU-no-action & $1.6312\pm0.0404$ & $9.886\pm0.143$ & $1.4432\pm0.1699$ & $11.863\pm0.153$ & $1.5290\pm0.2186$ & $12.682\pm0.297$\\
GRU-action & $1.6280\pm0.0353$ & $9.884\pm0.147$ & $1.4362\pm0.1627$ & $11.856\pm0.180$ & $1.5195\pm0.2139$ & $12.633\pm0.312$\\
Mask-decay GRU-action & $1.6556\pm0.1160$ & $9.806\pm0.527$ & $2.5722\pm0.8519$ & $15.091\pm4.204$ & $3.0011\pm0.7588$ & $17.324\pm4.258$\\
Transformer-no-action & $1.0762\pm0.0013$ & $9.038\pm0.004$ & $0.6149\pm0.0028$ & $10.040\pm0.012$ & $0.7906\pm0.0068$ & $10.965\pm0.023$\\
Transformer-action & $1.0750\pm0.0010$ & $9.034\pm0.011$ & $0.6107\pm0.0027$ & $10.008\pm0.011$ & $0.7812\pm0.0062$ & $10.899\pm0.021$\\
TFT-no-dynamic-action & $1.0655\pm0.0012$ & $8.994\pm0.022$ & $0.5996\pm0.0015$ & $9.999\pm0.015$ & $0.7234\pm0.0020$ & $10.747\pm0.013$\\
TFT-action & $1.0635\pm0.0023$ & $8.995\pm0.027$ & $\second{0.5937\pm0.0009}$ & $\second{9.966\pm0.023}$ & $\second{0.7114\pm0.0008}$ & $10.660\pm0.020$\\
\tftlarge{} & $\second{1.0622\pm0.0003}$ & $8.997\pm0.014$ & $0.5940\pm0.0008$ & $9.975\pm0.014$ & $0.7131\pm0.0019$ & $\second{10.652\pm0.029}$\\
\tftobj{} & $1.0920\pm0.0021$ & $\second{8.942\pm0.016}$ & $0.6112\pm0.0019$ & $10.127\pm0.062$ & $0.7298\pm0.0028$ & $10.796\pm0.055$\\
\midrule
\textbf{\method{} (ours)} & $\best{1.0620\pm0.0008}$ & $\best{8.883\pm0.005}$ & $\best{0.5924\pm0.0016}$ & $\best{9.894\pm0.005}$ & $\best{0.7106\pm0.0016}$ & $\best{10.613\pm0.015}$\\
\bottomrule
\end{tabular}
\caption{Descriptive all-valid terminal-step forecasting under the unified evaluator, reported as window-level mean $\pm$ seed SD within each horizon. Endpoint MSE is the unweighted 27-variable normalized MSE at the requested terminal hour, not a path-average metric. Test sizes are 6,046, 6,045, and 3,535 stays. Boldface and underlining mark the best and second-best values; inferential conclusions use stay-paired hierarchical intervals rather than rank alone.}
\label{tab:main-forecast}
\end{table*}

\begin{table*}[!t]
\centering
\scriptsize
\setlength{\tabcolsep}{3.1pt}
\resizebox{\textwidth}{!}{%
\begin{tabular}{lcccccc}
\toprule
& \multicolumn{2}{c}{8 hours} & \multicolumn{2}{c}{24 hours} & \multicolumn{2}{c}{48 hours}\\
\cmidrule(lr){2-3}\cmidrule(lr){4-5}\cmidrule(lr){6-7}
Model & Obs. endpoint MSE $\downarrow$ & Obs. MAP MAE $\downarrow$ & Obs. endpoint MSE $\downarrow$ & Obs. MAP MAE $\downarrow$ & Obs. endpoint MSE $\downarrow$ & Obs. MAP MAE $\downarrow$\\
\midrule
Mask-decay GRU-action & $1.4973\pm0.0466$ & $9.889\pm0.551$ & $1.9416\pm0.3011$ & $15.292\pm4.351$ & $2.2236\pm0.2503$ & $17.638\pm4.397$\\
TFT-action & $1.3059\pm0.0016$ & $9.075\pm0.025$ & $\best{1.0408\pm0.0012}$ & $\second{9.993\pm0.020}$ & $\best{1.0567\pm0.0003}$ & $10.656\pm0.023$\\
\tftlarge{} & $\second{1.3058\pm0.0004}$ & $9.080\pm0.014$ & $1.0415\pm0.0008$ & $10.001\pm0.017$ & $1.0584\pm0.0005$ & $\second{10.650\pm0.030}$\\
\tftobj{} & $1.3196\pm0.0011$ & $\second{9.029\pm0.011}$ & $1.0568\pm0.0003$ & $10.143\pm0.056$ & $1.0728\pm0.0008$ & $10.777\pm0.053$\\
\textbf{\method{}} & $\best{1.3055\pm0.0008}$ & $\best{8.976\pm0.002}$ & $\second{1.0411\pm0.0011}$ & $\best{9.928\pm0.004}$ & $\second{1.0577\pm0.0003}$ & $\best{10.609\pm0.019}$\\
\bottomrule
\end{tabular}
}%
\caption{MIMIC-IV target-observed-only evaluation, mean $\pm$ seed SD. Only terminal entries measured at the target hour contribute, and all models use identical entries. MAP MAE is in mmHg. Boldface and underlining mark the best and second-best values within the displayed comparison roster; rank does not imply pairwise significance.}
\label{tab:observed-forecast}
\end{table*}

\subsection{Cross-Database Paired MAP Analyses}
\label{app:cross-database-paired}

Table~\ref{tab:cross-database-paired} reports the paired MAP analyses summarized in the main text. Negative factual differences favor \method{}. Positive difference-in-gap values indicate that replacing the recorded action path causes greater error degradation for \method{} than for the comparator. These estimands use stay-level paired hierarchical inference rather than direct subtraction of the descriptive window-level means.

\begin{table*}[!t]
\centering
\scriptsize
\setlength{\tabcolsep}{2.4pt}
\resizebox{\textwidth}{!}{%
\begin{tabular}{llcccc}
\toprule
\multicolumn{6}{l}{\textbf{Forecasting differences in MAP MAE (mmHg), 95\% paired bootstrap CI}}\\
Dataset & Comparator & Terminal MAP & Observed terminal MAP & Integrated 1--48 h MAP & Integrated observed 1--48 h MAP\\
\midrule
MIMIC-IV & TFT-action
& $-0.0667$ [$-0.0828,-0.0478$]
& $-0.0512$ [$-0.0688,-0.0314$]
& $-0.0776$ [$-0.1011,-0.0536$]
& $-0.0713$ [$-0.0957,-0.0462$]\\

MIMIC-IV & \tftlarge{}
& $-0.0634$ [$-0.0801,-0.0483$]
& $-0.0470$ [$-0.0637,-0.0334$]
& $-0.0844$ [$-0.1208,-0.0521$]
& $-0.0794$ [$-0.1134,-0.0448$]\\

\midrule
eICU & TFT-action
& $-0.0412$ [$-0.0514,-0.0307$]
& $-0.0378$ [$-0.0483,-0.0275$]
& $-0.0442$ [$-0.0582,-0.0298$]
& $-0.0389$ [$-0.0543,-0.0255$]\\

eICU & \tftlarge{}
& $-0.0422$ [$-0.0594,-0.0239$]
& $-0.0399$ [$-0.0553,-0.0249$]
& $-0.0418$ [$-0.0616,-0.0218$]
& $-0.0382$ [$-0.0578,-0.0180$]\\

\addlinespace[3pt]
\midrule
\multicolumn{6}{l}{\textbf{Difference-in-gap in MAP MAE (mmHg), 95\% paired bootstrap CI}}\\
Dataset & Comparator & Full stream & Future only & Complete-stream change & Future-path change at 8/24/48 h\\
\midrule
MIMIC-IV & TFT-action
& $+0.0264$ [$+0.0158,+0.0408$]
& $+0.0255$ [$+0.0163,+0.0387$]
& 22.37\%
& 13.33 / 20.64 / 31.04\%\\

MIMIC-IV & \tftlarge{}
& $+0.0312$ [$+0.0225,+0.0412$]
& $+0.0322$ [$+0.0231,+0.0437$]
& 22.37\%
& 13.33 / 20.64 / 31.04\%\\

\midrule
eICU & TFT-action
& $+0.0166$ [$+0.0078,+0.0286$]
& $+0.0152$ [$+0.0058,+0.0282$]
& 16.62--16.81\%
& mean 11.2 / 17.6 / 26.8\%\\

eICU & \tftlarge{}
& $+0.0178$ [$+0.0030,+0.0289$]
& $+0.0181$ [$+0.0015,+0.0307$]
& 16.62--16.81\%
& mean 11.2 / 17.6 / 26.8\%\\
\bottomrule
\end{tabular}%
}
\caption{Cross-database paired MAP forecasting and action-path analyses. The upper block reports \method{}-minus-comparator factual MAP differences; negative values favor \method{}. The lower block reports differences in shifted-minus-recorded action gaps; positive values indicate stronger action-path dependence for \method{}. Path-change percentages quantify how often donor substitution changes the corresponding supplied action sequence. The eight prespecified eICU comparisons remain significant after correction within their respective hypothesis families.}
\label{tab:cross-database-paired}
\end{table*}

\newpage
\subsection{Evaluation Cohorts and Reporting Conventions}
\label{app:cohorts}

Benchmark tables report window-level means within each seed. Paired inference instead averages windows within stay and weights reporting horizons equally. A stay-macro difference therefore need not equal the difference between two displayed means.

The five stress-test groups overlap. ``Path treated'' means at least one action occurs in the candidate path. ``Action initiation'' means action is absent at the index boundary and appears within the future path. MAP deterioration denotes a raw MAP decrease of at least 10 mmHg. The locked shock-like group requires $[(\Delta\mathrm{MAP}\leq-10)\lor(\mathrm{MAP}_{t+h}<65)]$ and $[(\Delta\mathrm{lactate}\geq1)\lor(\mathrm{lactate}_{t+h}\geq2)]$. Group membership is used only for evaluation.

\begin{table*}[!t]
\centering
\small
\setlength{\tabcolsep}{4pt}
\begin{tabular}{llrr}
\toprule
Horizon & Group & Stays & Windows\\
\midrule
8 & All valid & 6,046 & 280,438\\
8 & Path treated & 673 & 18,694\\
8 & Action initiation & 673 & 5,524\\
8 & MAP deterioration & 5,928 & 58,934\\
8 & Shock-like deterioration & 1,274 & 11,885\\
\midrule
24 & All valid & 6,045 & 183,702\\
24 & Path treated & 673 & 18,958\\
24 & Action initiation & 673 & 9,799\\
24 & MAP deterioration & 5,226 & 42,041\\
24 & Shock-like deterioration & 890 & 7,604\\
\midrule
48 & All valid & 3,535 & 69,703\\
48 & Path treated & 560 & 10,819\\
48 & Action initiation & 560 & 7,743\\
48 & MAP deterioration & 2,856 & 16,606\\
48 & Shock-like deterioration & 418 & 2,679\\
\bottomrule
\end{tabular}
\caption{Final test-set stay and window counts. Groups overlap and do not partition the cohort.}
\label{tab:group-counts}
\end{table*}

\section{Cohort Construction, Preprocessing, and Exposure}
\label{app:cohort-details}

\subsection{Cohort Construction and Characteristics}
\label{app:cohort-flow}

MIMIC-IV v3.1 is a de-identified critical-care database distributed through PhysioNet \citep{johnson2023mimiciv,johnson2024mimiciv31}. The analysis uses only released database fields and the fixed cohort rules below.

\begin{table*}[!t]
\centering
\small
\begin{tabular}{lr}
\toprule
Stage & Stays\\
\midrule
All MIMIC-IV ICU stays & 94,458\\
Valid ICU timestamps & 94,444\\
Adult with valid timestamps & 94,444\\
Adult, valid timestamps, ICU LOS $\geq$24h & 74,829\\
Chronologically first stay satisfying basic criteria per subject & 54,551\\
Existing cohort\_base artifact rows & 54,551\\
\quad Excluded: pre-index vasopressor in ICU hours $[0,6)$ only & 14,243\\
\quad Excluded: death at or before ICU hour 6 only & 4\\
\quad Excluded: both pre-index exposure and death by hour 6 & 3\\
Final index-eligible roster & 40,301\\
Final tensorized cohort (no additional exclusions) & 40,301\\
Training split & 28,210\\
Validation split & 6,045\\
Test split & 6,046\\
\bottomrule
\end{tabular}
\caption{Cohort construction. The three exclusion rows are mutually exclusive and sum to 14,250. The 40,301-stay final index-eligible roster exactly matches the final tensorized cohort used for training, validation, and evaluation.}
\label{tab:cohort-flow}
\end{table*}

\begin{table*}[!t]
\centering
\small
\setlength{\tabcolsep}{4pt}
\begin{tabular}{lrrrrrr}
\toprule
Split & $N$ & Age, mean $\pm$ SD & Female (\%) & Mortality (\%) & LOS mean (d) & LOS median (d)\\
\midrule
Overall & 40,301 & 64.1 $\pm$ 17.4 & 44.7 & 9.6 & 4.10 & 2.31\\
Training & 28,210 & 64.1 $\pm$ 17.3 & 44.8 & 9.6 & 4.10 & 2.31\\
Validation & 6,045 & 63.8 $\pm$ 17.2 & 44.7 & 9.6 & 4.07 & 2.30\\
Test & 6,046 & 64.3 $\pm$ 17.6 & 44.3 & 9.6 & 4.12 & 2.33\\
\bottomrule
\end{tabular}
\caption{Cohort characteristics. Mortality is in-hospital mortality; LOS is ICU length of stay.}
\label{tab:cohort-characteristics}
\end{table*}

\begin{table}[tb]
\centering
\small
\begin{tabular}{lr}
\toprule
Test demographic category & $n$\\
\midrule
Female / male & 2,678 / 3,368\\
White & 3,887\\
Black & 612\\
Asian & 184\\
Hispanic/Latino & 234\\
Other/Unknown & 1,129\\
\bottomrule
\end{tabular}
\caption{Test-set sex and harmonized race-group counts used for exploratory subgroup analyses.}
\label{tab:test-demographics}
\end{table}

\subsection{Cohort and Preprocessing Details}
\label{app:preprocessing}

Adults with valid ICU admission/discharge timestamps and at least 24 hours of ICU stay are eligible; the chronologically first basic-eligible stay per subject is retained, yielding 54,551 stays. The fixed index at ICU hour 6 then excludes 14,243 stays with vasopressor exposure during hours $[0,6)$, 4 stays with death at or before hour 6, and 3 satisfying both conditions. The remaining 40,301 stay IDs exactly match the final tensor artifact; no additional stay is removed during tensorization or subsequent preprocessing. Multiple values in the same feature-hour are averaged, Fahrenheit temperatures are converted to Celsius, and observation masks are created before imputation.

The 27 variables are heart rate, respiratory rate, oxygen saturation, systolic blood pressure, diastolic blood pressure, MAP, temperature, GCS eye, GCS verbal, GCS motor, albumin, anion gap, bicarbonate, total bilirubin, creatinine, glucose, sodium, potassium, chloride, blood urea nitrogen, lactate, hemoglobin, platelet count, white blood cell count, INR, prothrombin time, and partial thromboplastin time. Patient-disjoint splits are fixed before any imputation or scaling statistic is estimated. Values are forward-filled using only prior observations within each stay; remaining leading gaps use medians estimated exclusively from final-training stays. Normalization statistics are estimated from valid training hours and applied unchanged to validation and test data.

The action map includes dopamine, epinephrine, norepinephrine, phenylephrine, and vasopressin input events. Cancelled or rewritten records, nonpositive amount/rate records, and nonpositive-duration intervals are removed. Retained intervals are mapped by half-open overlap. Historical and future action-memory features use the same boundary convention. Each hourly action is represented by the four-dimensional feature vector $[a,o,d/24,r/24]$, where $o$ marks the first hour of an exposure episode, $d$ is the duration of the active episode, and $r$ is the time since the most recent exposure. At forecast lead $k$, $\mathbf u^{\mathrm{fut}}_{i,k}=\mathbf u_{i,t+k-1}$. The normalized duration and time-since features are clipped to $[0,3]$, corresponding to a maximum encoded interval of 72 hours. Historical temporal inputs comprise normalized absolute ICU hour and a validity indicator for padded positions. Future temporal inputs comprise relative forecast lead and absolute target ICU hour within the trajectory. Overall valid-hour exposure is 4.65\%; the test rate is 4.46\%.

\subsection{Observation, Carry-Forward, and Endpoint Measurement Rates}
\label{app:missingness}

\begin{table*}[!t]
\centering
\scriptsize
\setlength{\tabcolsep}{2.5pt}
\begin{tabular}{lrrrrrrr}
\toprule
Variable & Direct obs. & Carried & Median fallback & Median age (h) & Endpoint 8h & Endpoint 24h & Endpoint 48h\\
\midrule
DBP & 86.1 & 12.4 & 1.6 & 1 & 86.0 & 84.2 & 83.8\\
GCS eye & 30.9 & 67.0 & 2.1 & 2 & 29.7 & 28.2 & 28.0\\
GCS motor & 30.8 & 67.1 & 2.1 & 2 & 29.6 & 28.1 & 27.9\\
GCS verbal & 30.8 & 67.0 & 2.1 & 2 & 29.7 & 28.2 & 28.0\\
Heart rate & 92.9 & 5.9 & 1.2 & 1 & 93.4 & 92.4 & 92.5\\
MAP & 86.1 & 12.3 & 1.6 & 1 & 86.1 & 84.2 & 83.8\\
Respiratory rate & 91.5 & 6.9 & 1.6 & 1 & 92.0 & 90.8 & 90.9\\
SBP & 86.1 & 12.4 & 1.6 & 1 & 86.1 & 84.2 & 83.8\\
SpO$_2$ & 91.0 & 7.7 & 1.2 & 1 & 91.4 & 90.0 & 90.1\\
Temperature & 27.5 & 69.2 & 3.3 & 2 & 26.8 & 26.1 & 26.7\\
Albumin & 1.1 & 29.7 & 69.3 & 18 & 0.8 & 0.7 & 0.7\\
Anion gap & 7.1 & 83.3 & 9.6 & 7 & 6.3 & 6.0 & 5.9\\
Bicarbonate & 7.1 & 83.4 & 9.5 & 7 & 6.3 & 6.0 & 5.9\\
Bilirubin & 2.0 & 43.1 & 54.9 & 14 & 1.6 & 1.4 & 1.4\\
BUN & 7.1 & 83.5 & 9.4 & 7 & 6.3 & 6.0 & 5.9\\
Chloride & 7.4 & 83.3 & 9.3 & 7 & 6.6 & 6.3 & 6.2\\
Creatinine & 7.1 & 83.4 & 9.4 & 7 & 6.4 & 6.1 & 5.9\\
Glucose & 6.9 & 82.2 & 10.9 & 7 & 6.3 & 6.0 & 5.9\\
Hemoglobin & 6.8 & 83.0 & 10.3 & 8 & 5.9 & 5.5 & 5.4\\
INR & 4.6 & 74.2 & 21.2 & 11 & 3.7 & 3.4 & 3.3\\
Lactate & 3.7 & 46.4 & 50.0 & 13 & 2.8 & 2.2 & 2.1\\
Platelets & 6.8 & 83.1 & 10.2 & 8 & 5.8 & 5.5 & 5.3\\
Potassium & 7.5 & 82.7 & 9.8 & 7 & 6.8 & 6.5 & 6.4\\
PT & 4.6 & 74.2 & 21.2 & 11 & 3.7 & 3.4 & 3.3\\
PTT & 4.9 & 73.5 & 21.6 & 10 & 4.1 & 3.8 & 3.7\\
Sodium & 7.5 & 82.4 & 10.1 & 7 & 6.8 & 6.6 & 6.6\\
WBC & 6.7 & 83.1 & 10.2 & 8 & 5.8 & 5.5 & 5.3\\
\bottomrule
\end{tabular}
\caption{Test-set valid-hour preprocessing and endpoint observation rates, in percent except carry-forward age. ``Direct obs.'' is measurement in the hour; ``carried'' is a value forward-filled using only prior observations; fallback is the frozen training median. Endpoint columns give the fraction of all-valid windows in which the target variable was directly observed at the endpoint.}
\label{tab:missingness-full}
\end{table*}

Dense vital signs are measured frequently, but several laboratories are directly observed in fewer than 8\% of valid hours. Albumin, bilirubin, and lactate have especially large fallback fractions. Target-observed-only evaluation therefore complements rather than replaces the operational tensorized task.

\subsection{Action Sparsity and Drug Exposure}
\label{app:action-sparsity}

\begin{table*}[!t]
\centering
\scriptsize
\setlength{\tabcolsep}{3pt}
\begin{tabular}{lrrrrrrrr}
\toprule
Split & Zero-action stays & Positive valid hours & Path 8h & Path 24h & Path 48h & Initiation 8h & Initiation 24h & Initiation 48h\\
\midrule
Overall & 88.7 & 4.65 & 6.86 & 10.50 & 15.76 & 2.01 & 5.46 & 11.48\\
Test & 88.8 & 4.46 & 6.67 & 10.32 & 15.52 & 1.97 & 5.33 & 11.11\\
\bottomrule
\end{tabular}
\caption{Action sparsity percentages. Path and initiation fractions use all-valid windows at each horizon.}
\label{tab:action-sparsity}
\end{table*}

\begin{table}[tb]
\centering
\small
\begin{tabular}{lrr}
\toprule
Drug & Exposed stays & Cohort fraction (\%)\\
\midrule
Dopamine & 339 & 0.84\\
Epinephrine & 583 & 1.45\\
Norepinephrine & 2,835 & 7.03\\
Phenylephrine & 3,649 & 9.05\\
Vasopressin & 986 & 2.45\\
\bottomrule
\end{tabular}
\caption{Stay-level retained vasopressor exposure in the full 40,301-stay cohort. Categories overlap because a stay can receive multiple drugs.}
\label{tab:drug-exposure}
\end{table}

The exact audit shows that the 14,250 exclusions after first-stay selection comprise 14,243 pre-index vasopressor-only exclusions, 4 death-by-hour-6-only exclusions, and 3 satisfying both conditions. The 40,301 retained stay IDs exactly match the final tensor roster, so downstream tensorization and preprocessing introduce no additional stay exclusion.

\subsection{Action-Path Change Audit}
\label{app:action-change}

\begin{table*}[!t]
\centering
\scriptsize
\setlength{\tabcolsep}{3.0pt}
\begin{tabular}{lrrrrr}
\toprule
Protocol / horizon & Eligible units & Raw path changed (\%) & Action-memory changed (\%) & Original nonzero changed (\%) & Zero $\rightarrow$ zero (\%)\\
\midrule
Complete 72-hour stream & 6,045 & 22.37 & 22.37 & 100.0 & 77.63\\
Full stream, 8h & 280,422 & 13.33 & 20.70 & 100.0 & 86.67\\
Full stream, 24h & 183,702 & 20.64 & 25.34 & 100.0 & 79.36\\
Full stream, 48h & 69,703 & 31.04 & 32.61 & 100.0 & 68.96\\
Future only, 8h & 280,422 & 13.33 & 13.33 & 100.0 & 86.67\\
Future only, 24h & 183,702 & 20.64 & 20.64 & 100.0 & 79.36\\
Future only, 48h & 69,703 & 31.04 & 31.04 & 100.0 & 68.96\\
\bottomrule
\end{tabular}
\caption{Actual action changes induced by the five fixed derangements. Complete-stream percentages use matched test stays; horizon-specific percentages use shift-eligible all-valid windows. Raw future exposure is identical between the two protocols, while full-stream substitution can additionally change historical action-memory features.}
\label{tab:action-change}
\end{table*}

Each derangement is one-to-one and contains no self-donor among the 6,045 eligible test stays. Every replicate changes 1,352 complete streams, including all 676 originally nonzero streams; 4,693 assignments remain all-zero to all-zero. Although the per-replicate count is fixed by construction, donor identities differ: 2,985 stays change in at least one of the five derangements and 677 change in all five. Unchanged zero-to-zero units contribute no action perturbation and attenuate population-level all-valid gaps toward zero. The fine-grained analysis in Section~\ref{app:changed-window-audit} conditions on realized future-path changes rather than averaging these no-op assignments together with changed assignments.

\subsection{Treatment-Likelihood Static Context}
\label{app:propensity}

The two static features inherited from the TFT branch are predictive context rather than causal weights. A logistic regression is fitted only on final-training stays to predict any retained vasopressor exposure during hours $[6,12)$. Inputs comprise 55 numeric and 8 categorical pre-index summaries. Training values use five-fold out-of-fold predictions; validation and test use a model fitted on all final-training stays. The features are clipped $p_i$ and $\min(p_i,1-p_i)$. They do not enter inverse-probability weighting or donor assignment.

\section{Training and Direct-Model Controls}
\label{app:training-controls}

\subsection{Training Curriculum and Reproducibility}
\label{app:training-details}

\begin{table*}[!t]
\centering
\small
\setlength{\tabcolsep}{3.5pt}
\textbf{Panel A: stage-dependent objective weights}\\[2pt]
\begin{tabular}{lrrrrrrr}
\toprule
Stage & Epochs & Trajectory & Endpoint & Hemo & Anchor & Distill & Pre-exposure / Budget\\
\midrule
1 & 4 & 1.00 & 0.60 & 0.35 & 0.20 & 0.00 & 0.00 / 0.02\\
2 & 6 & 1.00 & 0.60 & 0.30 & 0.10 & 0.20 & 0.00 / 0.02\\
3 & 6 & 1.00 & 0.60 & 0.30 & 0.05 & 0.15 & 0.02 / 0.02\\
4 & 4 & 1.00 & 0.60 & 0.30 & 0.00 & 0.10 & 0.02 / 0.02\\
\bottomrule
\end{tabular}

\vspace{5pt}
\begin{minipage}[t]{0.39\textwidth}
\centering
\textbf{Panel B: trainable gate initializations}\\[2pt]
\begin{tabular}{lcc}
\toprule
Parameter & Initialization & Trainable\\
\midrule
$b_a^{(0)}$ & $-2$ & Yes\\
$b_c^{(0)}$ & $-3$ & Yes\\
$b_H^{(0)}$ & $-2.5$ & Yes\\
\bottomrule
\end{tabular}
\end{minipage}\hfill
\begin{minipage}[t]{0.58\textwidth}
\centering
\textbf{Panel C: frozen checkpoint-score coefficients}\\[2pt]
\begin{tabular}{lcl}
\toprule
Parameter & Value & Role\\
\midrule
$w_{\mathrm{end}}$ & 0.45 & DRIFT relative endpoint term\\
$w_{\mathrm{MAP}}$ & 0.55 & DRIFT relative MAP term\\
$v_{\mathrm{end}}$ & 0.80 & Direct-control endpoint term\\
$v_{\mathrm{MAP}}$ & 0.02 & Direct-control MAP term\\
\bottomrule
\end{tabular}
\end{minipage}
\caption{Training specification for \method{} and the direct controls. Panel B lists negative trainable gate-bias initializations, not fixed residual budgets. Panel C lists coefficients fixed before formal test evaluation; neither test outcomes nor action-replacement gaps enter checkpoint selection.}
\label{tab:stage-weights}
\end{table*}

The negative biases in Panel B initialize sigmoid gate values below 0.12. Together with zero-initialized output projections, they allow each new path to begin with little influence and increase its contribution only through training; they do not impose a fixed final gate value.

\paragraph{Complete training objective.}
Let $B$ be the minibatch size, $H\in\{8,16,24,48\}$ the horizon sampled for the minibatch, $D=27$, and $d_z=128$. The normalized prediction and target are $\hat{x}_{i,k,j}$ and $x_{i,k,j}$; $\hat{y}_{i,k,j}$ and $y_{i,k,j}$ denote their raw-scale counterparts. Training windows already satisfy the temporal-validity rules. Missing values are processed by the prespecified past-only within-stay filling pipeline, with fallback medians and scaling statistics estimated exclusively from the training split. The seven losses are computed on the resulting complete tensorized targets; the future observation mask is retained for the observed-target secondary evaluations in Section~A rather than applied to the primary training objective.

The feature-weighted trajectory and terminal losses are
\begin{align}
\mathcal L_{\mathrm{traj}}
&=
\frac{1}{BH}\sum_{i=1}^{B}\sum_{k=1}^{H}
\frac{\sum_{j=1}^{D}w_j(\hat{x}_{i,k,j}-x_{i,k,j})^2}
{\sum_{j=1}^{D}w_j},
\label{eq:supp-traj-loss}\\
\mathcal L_{\mathrm{end}}
&=
\frac{1}{BD}\sum_{i=1}^{B}\sum_{j=1}^{D}
(\hat{x}_{i,H,j}-x_{i,H,j})^2,
\label{eq:supp-end-loss}
\end{align}
where $w_j=2$ for MAP, lactate, and creatinine and $w_j=1$ otherwise. Thus, the terminal term is unweighted across channels even though the trajectory term is not.

For raw-scale hemodynamics, define
\begin{equation}
\ell_{\delta}(e)=
\begin{cases}
\tfrac12e^2, & |e|\leq\delta,\\
\delta(|e|-\tfrac12\delta), & |e|>\delta.
\end{cases}
\end{equation}
With $\mathcal H=\{\mathrm{HR},\mathrm{SBP},\mathrm{DBP},\mathrm{MAP}\}$, $\boldsymbol\nu=(0.20,0.25,0.25,1.00)$ in that order, and $\delta_H=5.0$ for all four channels,
\begin{equation}
\mathcal L_{\mathrm{hemo}}
=
\frac{1}{BH}\sum_{i=1}^{B}\sum_{k=1}^{H}
\frac{\sum_{r\in\mathcal H}\nu_r
\ell_{5}(\hat{y}_{i,k,r}-y_{i,k,r})}
{\sum_{r\in\mathcal H}\nu_r}.
\label{eq:supp-hemo-loss}
\end{equation}

Let $\hat{x}^{\mathrm{anchor}}_{i,k,j}$ be the frozen TFT-action prediction. Anchor preservation is imposed in prediction space,
\begin{equation}
\mathcal L_{\mathrm{anchor}}
=\frac{1}{BHD}\sum_{i=1}^{B}\sum_{k=1}^{H}\sum_{j=1}^{D}
\left[\hat{x}_{i,k,j}-\operatorname{sg}
(\hat{x}^{\mathrm{anchor}}_{i,k,j})\right]^2.
\label{eq:supp-anchor-loss}
\end{equation}
For valid-history indicator $v_{i,t}$ and historical observation mask $m_{i,t,j}$, the density weight and direct-to-recursive state distillation \citep{hinton2015distilling} are
\begin{align}
\rho_i
&=\operatorname{clip}\!\left(
\frac{\sum_t v_{i,t}(D^{-1}\sum_jm_{i,t,j})}
{\max(\sum_tv_{i,t},1)},0.1,1\right),\\
\mathcal L_{\mathrm{distill}}
&=\frac{1}{BH}\sum_{i=1}^{B}\sum_{k=1}^{H}
\frac{\rho_i}{d_z}
\left\|\zrec_{i,k}-\operatorname{sg}(\zdir_{i,k})\right\|_2^2.
\label{eq:supp-distill-loss}
\end{align}
The stop-gradient target is the adapted direct state, so this term updates only the recursive pathway.

Let $q_{i,k}$ indicate whether the supplied future action prefix through lead $k$ contains an exposure, and let $\boldsymbol\Delta^a_{i,k}$ be the bounded raw action-response latent before prefix activation and response gating. Pre-exposure suppression is
\begin{equation}
\mathcal L_{\mathrm{pre}}
=\frac{1}{BHd_z}\sum_{i=1}^{B}\sum_{k=1}^{H}
(1-q_{i,k})\|\boldsymbol\Delta^a_{i,k}\|_2^2.
\label{eq:supp-pre-loss}
\end{equation}
Finally, let $\boldsymbol\delta^s$ be the realized direct-state adapter output, $\mathbf c^{\mathrm{latent}}$ the gated discrepancy correction, and $\mathbf c^H=\mathbf c^{H,0}+\mathbf c^{H,a}$ the four-channel local correction. Correction-magnitude regularization is
\begin{equation}
\mathcal L_{\mathrm{budget}}
=\frac{\|\boldsymbol\delta^s\|_F^2}{BHd_z}
+\frac{\|\mathbf c^{\mathrm{latent}}\|_F^2}{BHd_z}
+\frac{\|\mathbf c^H\|_F^2}{4BH}.
\label{eq:supp-budget-loss}
\end{equation}
This is the sum of three elementwise mean-square penalties, not a thresholded budget-exceedance loss. The stage-specific weighted sum uses the coefficients in Table~\ref{tab:stage-weights}.

\paragraph{Recursive-state initialization.}
The recursive initialization adapter is
\begin{equation}
P_{\mathrm{init}}(\mathbf h_t)
=
\alpha_s\tanh\!\left(
P_{\mathrm{init},2}
\operatorname{GELU}\!\left(
P_{\mathrm{init},1}\operatorname{LN}(\mathbf h_t)
\right)\right),
\end{equation}
where $P_{\mathrm{init},1}$ and $P_{\mathrm{init},2}$
are affine layers forming a
$128\rightarrow32\rightarrow128$ bottleneck adapter.
The frozen maximum residual scale is $\alpha_s=0.10$.
The recursive state is initialized as
\begin{equation}
\zrec_0
=
\operatorname{LN}\!\left(
\mathbf h_t+P_{\mathrm{init}}(\mathbf h_t)
\right).
\end{equation}
Both the weight and bias of the final $32\rightarrow128$
affine layer are initialized to zero. Therefore,
$P_{\mathrm{init}}(\mathbf h_t)=\mathbf 0$ at initialization,
and $\zrec_0=\operatorname{LN}(\mathbf h_t)$ before the
adapter learns a nonzero residual. The recursive initializer
and the direct-state adapter have separate parameters, although
they use the same frozen default residual bound and are scaled
together in the residual-budget sensitivity analysis.
\paragraph{Correction and local-branch definitions.}
For the discrepancy $\mathbf d_{i,k}=\zrec_{i,k}-\zdir_{i,k}$, the scalar correction gate is
\begin{equation}
g^c_{i,k}=\sigma G_c\!\left([\operatorname{LN}(\mathbf d_{i,k}),\boldsymbol\tau_k,\mathbf s_i]\right).
\end{equation}
$G_c$ is a GELU MLP with hidden width 64 and a scalar output. Its output weights are initialized to zero and its bias to $b_c^{(0)}$; the $128\rightarrow32\rightarrow128$ discrepancy decoder also has a zero-initialized output projection.

For the four local channels, the learned decay \citep{che2018grud} and recurrent token are
\begin{align}
\boldsymbol\gamma_{i,t}&=\exp[-\operatorname{ReLU}(W_\delta\boldsymbol\delta^H_{i,t}+\mathbf b_\delta)],\\
\boldsymbol\xi^H_{i,t}&=[\boldsymbol\gamma_{i,t}\odot\mathbf x^H_{i,t},\mathbf m^H_{i,t},\boldsymbol\delta^H_{i,t}].
\end{align}
The token $\boldsymbol\xi^H_{i,t}$ is projected to 64 dimensions and processed by a one-layer 64-dimensional GRU \citep{cho2014gru}, whose final state is $\mathbf h_i^H$. Let $\mathbf v^H_{i,k}=[\mathbf h_i^H,\boldsymbol\tau_k,\mathbf s_i]$. The future-action-independent base and action corrections are
\begin{align}
g^{H,0}_{i,k}&=\sigma G_H(\mathbf v^H_{i,k}),\\
\mathbf c^{H,0}_{i,k}&=g^{H,0}_{i,k}\alpha_H\tanh F_H(\mathbf v^H_{i,k}),\\
\mathbf c^{H,a}_{i,k}&=\alpha_H\tanh(W_{H,a}\mathbf e^a_{i,k}).
\end{align}
$F_H$ is a $68\rightarrow64\rightarrow4$ GELU MLP. Its output layer and $W_{H,a}$ are zero initialized. The scalar gate $G_H$ has zero-initialized output weights and trainable initial bias $b_H^{(0)}$. Thus, future time and static context enter the local base, whereas only the effective recursive response $\mathbf e^a_{i,k}$ enters the local action term.

All neural results use seeds 42, 43, and 44. \method{} trains for all 20 scheduled epochs, with batch size 192 and 500 sampled minibatches per epoch. Horizons $\{8,16,24,48\}$ are sampled with probabilities $(0.30,0.15,0.30,0.25)$, one horizon per minibatch. AdamW \citep{loshchilov2019decoupled} is reinitialized at the beginning of each curriculum stage. New modules use learning rate $10^{-4}$ and weight decay $10^{-4}$; only the anchor pre-output block and output layer are unfrozen in Stage~4, with learning rate $10^{-5}$. Gradients are clipped to global norm 0.5. \method{} uses no learning-rate scheduler, warmup, or early stopping. Checkpoints are selected among the 20 trained epochs using factual validation forecasting only; validation action diagnostics never enter selection. Validation forecasting uses a deterministic cap of 50,000 windows per reporting horizon.

Training uses bfloat16 autocast, with loss calculations in float32. Training and evaluation runs enable TF32 for CUDA matrix multiplication and cuDNN operations. Random, NumPy, and PyTorch generators are initialized from each fixed seed, but cuDNN benchmarking is enabled and deterministic algorithms are not enforced; results should therefore not be interpreted as bitwise deterministic across hardware or software stacks.

\begin{table}[t]
\centering
\small
\setlength{\tabcolsep}{4pt}
\begin{tabular}{@{}p{0.27\columnwidth}p{0.66\columnwidth}@{}}
\toprule
Component & Version or configuration\\
\midrule
CPU & Intel Xeon Platinum 8480+\\
System memory & 3.9 TiB\\
GPU node & Eight NVIDIA H100 NVL GPUs, 95,830 MiB each\\
GPU per run & One NVIDIA H100 NVL\\
Python & 3.11.15\\
Operating system & Linux 5.15, x86\_64, glibc 2.35\\
PyTorch & 2.11.0+cu128\\
CUDA runtime & 12.8\\
cuDNN & 9.19.0\\
NVIDIA driver & 580.126.09\\
Precision & BF16 autocast; FP32 losses; TF32 enabled\\
Optimizer & AdamW; global gradient clip 0.5\\
Seeds & 42, 43, and 44\\
\bottomrule
\end{tabular}
\caption{Reproducibility environment. Independent model--seed runs were parallelized across an eight-GPU node, but each run used one GPU.}
\label{tab:environment}
\end{table}

Training progressively activates the added pathways. Stage~1 trains the direct-state residual adapter and action-independent mask-decay hemodynamic branch. Stage~2 additionally trains the recursive controlled transition and low-rank discrepancy correction. Stage~3 activates the local action-response projection. Stage~4 unfreezes only the anchor pre-output block and output layer for low-rate adjustment; all other anchor parameters remain frozen.

\paragraph{Bounded residual budgets and sensitivity.}
The default budgets were fixed during primary-cohort development and transferred unchanged to the sealed eICU replication. The trainable gate-bias initializations $b_a^{(0)}$, $b_c^{(0)}$, and $b_H^{(0)}$ are listed in Table~\ref{tab:stage-weights}; they are not fixed residual bounds. Let $\boldsymbol\alpha^{(0)}$ collect the five frozen defaults in Panel A of Table~\ref{tab:residual-budget-sensitivity}. We evaluated sensitivity to their common scale using
\begin{equation}
\boldsymbol\alpha(m)=m\boldsymbol\alpha^{(0)}.
\end{equation}
We evaluated multipliers of $0.5\times$, $1.0\times$, and $2.0\times$. The frozen $1.0\times$ checkpoints and TFT-action outputs were reused; only the $0.5\times$ and $2.0\times$ variants were newly trained, with three seeds each. The sensitivity workflow did not read or modify any eICU artifact, checkpoint, or result.

\begin{table*}[!t]
\centering
\scriptsize
\setlength{\tabcolsep}{3.0pt}
\renewcommand{\arraystretch}{1.10}
\textbf{Panel A: frozen default residual budgets}\\[2pt]
\begin{tabular}{lcl}
\toprule
Symbol & Default & Role \\
\midrule
$\alpha_s$ & 0.10 & Direct-state and recursive-initial-state adapter budget \\
$\alpha_0$ & 0.20 & Background recursive-transition budget \\
$\alpha_a$ & 0.20 & Action-response recursive-transition budget \\
$\alpha_c$ & 0.25 & Final discrepancy-correction budget \\
$\alpha_H$ & 0.35 & Local hemodynamic base/action output budget \\
\bottomrule
\end{tabular}

\vspace{5pt}
\textbf{Panel B1: forecasting robustness, mean $\pm$ seed SD}\\[2pt]
\begin{tabular}{@{}lcccccc@{}}
\toprule
Scale & \shortstack{Endpoint\\MSE $\downarrow$} & \shortstack{Terminal\\MAP $\downarrow$} & \shortstack{Observed\\MAP $\downarrow$} & \shortstack{Integrated\\MAP $\downarrow$} & \shortstack{Integrated\\observed MAP $\downarrow$} & \shortstack{$\Delta$ terminal MAP vs TFT-action\\95\% CI $\downarrow$} \\
\midrule
$0.5\times$ & $0.788399\pm0.000855$ & $9.8024\pm0.0114$ & $9.8400\pm0.0120$ & $9.8949\pm0.0120$ & $9.9458\pm0.0110$ & $-0.06391$ [$-0.07925,-0.04739$] \\
$1.0\times$ (frozen) & $0.788348\pm0.000889$ & $9.7968\pm0.0070$ & $9.8376\pm0.0075$ & $9.8894\pm0.0092$ & $9.9438\pm0.0073$ & $-0.0667$ [$-0.0828,-0.0478$] \\
$2.0\times$ & $0.788248\pm0.000891$ & $9.7911\pm0.0077$ & $9.8319\pm0.0091$ & $9.8806\pm0.0092$ & $9.9360\pm0.0091$ & $-0.07029$ [$-0.08699,-0.05306$] \\
\bottomrule
\end{tabular}

\vspace{5pt}
\textbf{Panel B2: action dependence and correction utilization, mean $\pm$ seed SD}\\[2pt]
\resizebox{\textwidth}{!}{%
\begin{tabular}{@{}lcccccc@{}}
\toprule
Scale & \shortstack{Full-stream\\gap $\uparrow$} & \shortstack{Future-only\\gap $\uparrow$} & \shortstack{$\Delta$ full gap vs TFT-action\\95\% CI $\uparrow$} & \shortstack{$\Delta$ future gap vs TFT-action\\95\% CI $\uparrow$} & Latent RMS & Output RMS\\
\midrule
$0.5\times$ &
$0.1267\pm0.0029$ &
$0.1359\pm0.0045$ &
\shortstack{$+0.02934$\\$[+0.02160,+0.03611]$} &
\shortstack{$+0.02964$\\$[+0.02264,+0.03529]$} &
$0.01157\pm0.00381$ &
$0.02168\pm0.00142$\\
$1.0\times$ (frozen) &
$0.1237\pm0.0056$ &
$0.1318\pm0.0084$ &
\shortstack{$+0.02635$\\$[+0.01577,+0.04081]$} &
\shortstack{$+0.02545$\\$[+0.01626,+0.03872]$} &
$0.01011\pm0.00128$ &
$0.02653\pm0.00069$\\
$2.0\times$ &
$0.1317\pm0.0057$ &
$0.1411\pm0.0048$ &
\shortstack{$+0.03442$\\$[+0.02722,+0.04391]$} &
\shortstack{$+0.03476$\\$[+0.02827,+0.04337]$} &
$0.00773\pm0.00200$ &
$0.02881\pm0.00223$\\
\bottomrule
\end{tabular}%
}
\caption{Residual-budget definitions and additional MIMIC-IV test-set sensitivity analysis. Terminal forecasting metrics and action gaps are equally averaged over the 8-, 24-, and 48-hour reporting horizons; integrated metrics average leads 1--48 on the fixed 48-hour-eligible cohort. MAP quantities and gaps are in mmHg. In Panel B2, gap columns are shifted-minus-recorded values for \method{}, whereas $\Delta$ columns subtract the corresponding TFT-action gap. Panel B1 reports paired intervals for the terminal MAP difference, and Panel B2 reports paired intervals for both difference-in-gap columns. All evaluated scales retain lower MAP error and positive difference-in-gap values relative to TFT-action. This analysis was not used for configuration or checkpoint selection, and the frozen $1.0\times$ model remains primary.}
\label{tab:residual-budget-sensitivity}
\end{table*}

For final \method{}, trained epochs must pass a six-cell factual gate formed by endpoint MSE and MAP MAE at 8, 24, and 48 hours. Relative to the seed-matched anchor, no cell may degrade by more than $0.20\%$, and the mean relative change across horizons must favor the trained model for each metric. Among eligible epochs, selection minimizes
\begin{equation}
\mathcal S_{\mathrm{DRIFT}}
=w_{\mathrm{end}}\overline r_{\mathrm{end}}
+w_{\mathrm{MAP}}\overline r_{\mathrm{MAP}}.
\end{equation}
Here, the bars denote horizon averages of relative changes and the frozen coefficients are listed in Table~\ref{tab:stage-weights}.
The exact seed-matched TFT anchor is retained only as a safety fallback; reported \method{} checkpoints are trained epochs. Every retrained final-architecture ablation uses the same fixed factual composite among trained epochs, and epoch zero is prohibited.

TFT-Large and \tftobj{} use a deterministic validation cap of 60,000 windows per reporting horizon and select the epoch minimizing
\begin{equation}
\begin{aligned}
\ell_h
&=v_{\mathrm{end}}\mathrm{MSE}_{\mathrm{end}}(h)
+v_{\mathrm{MAP}}\mathrm{MAE}_{\mathrm{MAP}}(h),\\
\mathcal S_{\mathrm{direct}}
&=\frac{1}{3}\sum_{h\in\{8,24,48\}}\ell_h.
\end{aligned}
\end{equation}

Both checkpoint scores and their coefficients were fixed before formal test evaluation; no separate coefficient sweep was performed. The \method{} score uses relative changes because adaptation starts from a seed-matched TFT anchor and the factual gate measures preservation relative to that anchor. The direct controls have no corresponding anchor reference and therefore use an absolute validation composite. Action-replacement gaps and test-set outcomes never enter checkpoint selection.

\tftobj{} is trained from scratch with hidden size 140, four heads, one recurrent layer, batch size 192, at most 18 epochs, 1,000 sampled minibatches per epoch, and initial learning rate $2\times10^{-4}$. Its transferred forecast loss is weighted path MSE $+0.60$ terminal endpoint MSE $+0.30$ raw hemodynamic Huber, corresponding to the dominant later-stage forecast objective rather than the full stage-dependent curriculum. Selected epochs are 15, 12, and 13; early stopping uses patience five. No separate loss-weight sweep was performed for this control.

\subsection{Capacity-Matched and Forecast-Loss-Transfer Controls}
\label{app:controls}

\begin{table}[tb]
\centering
\small
\begin{tabular}{lrr}
\toprule
Model & Parameters & Difference from DRIFT\\
\midrule
TFT-action & 1,328,767 & $-14.35\%$\\
\method{} & 1,551,342 & 0\\
\tftlarge{} & 1,586,047 & $+2.24\%$\\
\tftobj{} & 1,586,047 & $+2.24\%$\\
\bottomrule
\end{tabular}
\caption{Inference-active parameter counts.}
\label{tab:params}
\end{table}

\begin{table*}[!t]
\centering
\scriptsize
\setlength{\tabcolsep}{3.0pt}
\begin{tabular}{llrrr}
\toprule
Evaluation & Comparator & Metric & Relative change & Mean difference [95\% CI]\\
\midrule
Standard & TFT-action & Endpoint MSE & $-0.154\%$ & $-0.001215$ [$-0.002718,-0.000212$]\\
Standard & TFT-action & MAP MAE & $-0.673\%$ & $-0.0667$ [$-0.0828,-0.0478$]\\
Standard & TFT-Large & Endpoint MSE & $-0.172\%$ & $-0.001359$ [$-0.003078,0.000510$]\\
Standard & TFT-Large & MAP MAE & $-0.639\%$ & $-0.0634$ [$-0.0801,-0.0483$]\\
Standard & TFT-Large-FL & Endpoint MSE & $-2.567\%$ & $-0.02079$ [$-0.02284,-0.01857$]\\
Standard & TFT-Large-FL & MAP MAE & $-1.296\%$ & $-0.1294$ [$-0.1723,-0.0924$]\\
\midrule
Observed & TFT-action & Endpoint MSE & $+0.028\%$ & $+0.000312$ [$-0.000371,0.000811$]\\
Observed & TFT-action & MAP MAE & $-0.514\%$ & $-0.0512$ [$-0.0688,-0.0314$]\\
Observed & TFT-Large & Endpoint MSE & $-0.003\%$ & $-0.000038$ [$-0.000902,0.000782$]\\
Observed & TFT-Large & MAP MAE & $-0.472\%$ & $-0.0470$ [$-0.0637,-0.0334$]\\
Observed & TFT-Large-FL & Endpoint MSE & $-1.083\%$ & $-0.01216$ [$-0.01307,-0.01127$]\\
Observed & TFT-Large-FL & MAP MAE & $-1.160\%$ & $-0.1163$ [$-0.1558,-0.0826$]\\
\bottomrule
\end{tabular}
\caption{Hierarchical macro comparisons of DRIFT against the three direct controls. Negative values favor DRIFT. All rows except the two observed multivariate comparisons and the standard TFT-Large multivariate comparison have $p_{\mathrm{boot}}\leq0.001$; those three have $p_{\mathrm{boot}}=0.361$, $0.963$, and $0.143$, respectively.}
\label{tab:control-comparisons}
\end{table*}

\subsection{Unified Checkpoint-Selection Robustness}
\label{app:checkpoint-robustness}

The original model classes used different factual validation scores for checkpoint selection. We therefore conducted a separate MIMIC-IV robustness experiment in which \method{} and TFT-action were retrained for seeds 42, 43, and 44 while retaining every validation-epoch checkpoint. Before test inference, one checkpoint per model and seed was selected and frozen under each of three shared validation criteria.

For each model--seed run, validation endpoint MSE and validation MAP MAE were independently min--max normalized over the retained epochs. The three scores were
\begin{align}
S_{\mathrm{endpoint}}
&=
0.80\,\widetilde{\mathrm{MSE}}_{\mathrm{endpoint}}
+
0.20\,\widetilde{\mathrm{MAE}}_{\mathrm{MAP}},\\
S_{\mathrm{equal}}
&=
0.50\,\widetilde{\mathrm{MSE}}_{\mathrm{endpoint}}
+
0.50\,\widetilde{\mathrm{MAE}}_{\mathrm{MAP}},\\
S_{\mathrm{MAP}}
&=
0.20\,\widetilde{\mathrm{MSE}}_{\mathrm{endpoint}}
+
0.80\,\widetilde{\mathrm{MAE}}_{\mathrm{MAP}}.
\end{align}
Lower scores are preferred. Criterion definitions, selected epochs, checkpoint hashes, data identities, donor mappings, and evaluator configuration were frozen before test inference. When multiple criteria selected the same checkpoint, that checkpoint was evaluated once and its output was reused without changing the estimand. The MIMIC-IV test roster, feature order, horizons, donor mappings, changed-window definitions, hierarchical bootstrap, and multiplicity correction were otherwise unchanged.

\begin{table*}[!t]
\centering
\small
\setlength{\tabcolsep}{4.0pt}
\begin{tabular}{lccc}
\toprule
Criterion &
8 h difference [95\% CI] &
24 h difference [95\% CI] &
48 h difference [95\% CI]\\
\midrule
Endpoint-heavy
& $-0.104$ [$-0.127,-0.081$]
& $-0.054$ [$-0.080,-0.031$]
& $-0.038$ [$-0.065,-0.011$]\\
Equal
& $-0.087$ [$-0.103,-0.072$]
& $-0.048$ [$-0.071,-0.024$]
& $-0.042$ [$-0.079,-0.012$]\\
MAP-heavy
& $-0.098$ [$-0.123,-0.073$]
& $-0.047$ [$-0.067,-0.025$]
& $-0.037$ [$-0.072,-0.010$]\\
\bottomrule
\end{tabular}
\caption{MIMIC-IV MAP robustness under shared validation checkpoint-selection criteria. Entries are \method{} minus TFT-action MAP MAE in mmHg with 95\% paired bootstrap intervals; negative values favor \method{}. All nine intervals exclude zero.}
\label{tab:checkpoint-map-robustness}
\end{table*}

Across all three shared criteria, \method{} retains lower MAP MAE than TFT-action at every horizon (Table~\ref{tab:checkpoint-map-robustness}). The advantage ranges from 0.087--0.104 mmHg at 8 hours, 0.047--0.054 mmHg at 24 hours, and 0.037--0.042 mmHg at 48 hours. All nine paired confidence intervals exclude zero. Thus, the primary MAP conclusion is not attributable to the original model-specific validation weighting.

We next applied the same changed-window estimands to the checkpoints frozen under each shared criterion. Benjamini--Hochberg correction was applied separately within two nine-comparison families: the changed-window factual family comprised three checkpoint criteria $\times$ three horizons, and the changed-window difference-in-gap family comprised three checkpoint criteria $\times$ three horizons. The two families were not pooled.

\begin{table*}[!t]
\centering
\scriptsize
\setlength{\tabcolsep}{2.4pt}
\renewcommand{\arraystretch}{1.12}
\resizebox{\textwidth}{!}{%
\begin{tabular}{@{}llccc@{}}
\toprule
Checkpoint criterion
& Estimand
& 8 h
& 24 h
& 48 h \\
\midrule

Endpoint-heavy
& Factual difference
& \shortstack{$-0.0989$\\$[-0.1236,-0.0758]$\\$(q=0.0020)$}
& \shortstack{$-0.0550$\\$[-0.0844,-0.0279]$\\$(q=0.0020)$}
& \shortstack{$-0.0384$\\$[-0.0735,-0.0031]$\\$(q=0.0396)$}
\\

Endpoint-heavy
& Difference-in-gap
& \shortstack{$+0.2389$\\$[+0.1711,+0.3000]$\\$(q=0.0011)$}
& \shortstack{$+0.2332$\\$[+0.0929,+0.3296]$\\$(q=0.0011)$}
& \shortstack{$+0.1910$\\$[+0.0960,+0.2712]$\\$(q=0.0011)$}
\\

\addlinespace[2pt]

Equal
& Factual difference
& \shortstack{$-0.0831$\\$[-0.1080,-0.0613]$\\$(q=0.0020)$}
& \shortstack{$-0.0535$\\$[-0.0891,-0.0215]$\\$(q=0.0034)$}
& \shortstack{$-0.0416$\\$[-0.0829,-0.0029]$\\$(q=0.0425)$}
\\

Equal
& Difference-in-gap
& \shortstack{$+0.2739$\\$[+0.2334,+0.3122]$\\$(q=0.0011)$}
& \shortstack{$+0.2756$\\$[+0.1996,+0.3380]$\\$(q=0.0011)$}
& \shortstack{$+0.2244$\\$[+0.1490,+0.3003]$\\$(q=0.0011)$}
\\

\addlinespace[2pt]

MAP-heavy
& Factual difference
& \shortstack{$-0.0922$\\$[-0.1181,-0.0650]$\\$(q=0.0020)$}
& \shortstack{$-0.0454$\\$[-0.0824,-0.0134]$\\$(q=0.0120)$}
& \shortstack{$-0.0378$\\$[-0.0789,-0.0041]$\\$(q=0.0360)$}
\\

MAP-heavy
& Difference-in-gap
& \shortstack{$+0.2470$\\$[+0.1723,+0.3121]$\\$(q=0.0011)$}
& \shortstack{$+0.2313$\\$[+0.0673,+0.3401]$\\$(q=0.0011)$}
& \shortstack{$+0.1793$\\$[+0.0349,+0.2967]$\\$(q=0.0120)$}
\\

\bottomrule
\end{tabular}%
}
\caption{Checkpoint-selection robustness in forecast windows where donor substitution changed at least one supplied future exposure. Factual differences are \method{} minus TFT-action observed-target MAP MAE under the recorded future action path; negative values favor \method{}. Difference-in-gap values are \method{} minus TFT-action differences between shifted-minus-recorded observed-target MAP error gaps; positive values indicate stronger conditional action-path dependence for \method{}. Brackets report 95\% confidence intervals from the stay-clustered, seed-aware hierarchical bootstrap. The reported $q$ values use Benjamini--Hochberg correction separately within the nine-comparison changed-window factual family and the nine-comparison changed-window difference-in-gap family.}
\label{tab:checkpoint-changed-window-robustness}
\end{table*}

Across all three shared checkpoint-selection criteria, the changed-window factual observed-target MAP difference favors \method{} at every horizon. All nine factual confidence intervals exclude zero, and all nine corresponding BH-adjusted $q$ values are below 0.05. The changed-window difference-in-gap is likewise positive in all nine criterion--horizon cells; every confidence interval excludes zero, and every adjusted $q$ value is below 0.05. Thus, selecting both models under endpoint-heavy, equal, or MAP-heavy validation weighting does not eliminate either \method{}'s factual MAP advantage in changed windows or its stronger conditional dependence on the supplied recorded future action path.

These results arise from the separate retraining and shared-selection experiment and do not replace the original frozen-checkpoint changed-window analysis in Table~\ref{tab:changed-window-map}. In that original analysis, the 48-hour factual point estimate favors \method{} but its confidence interval includes zero; under each of the three shared checkpoint-selection criteria in the separate robustness experiment, the 48-hour factual interval excludes zero. The 27-variable normalized-MSE direction is generally favorable but is not uniformly significant at 8 hours under equal and MAP-heavy selection. The robustness conclusion is therefore specific to MAP forecasting and changed-window action-path dependence rather than uniform superiority across every endpoint.

\subsection{Strengthened Action-Conditioned DirRecMO-8 Comparator}
\label{app:dirrecmo8}

\paragraph{Comparator architecture.}
We implemented a literature-inspired, action-conditioned DirRecMO-8 comparator to represent a conventional blockwise direct--recursive forecasting strategy \citep{lin2023pronet,green2025stratify}. The 48-hour forecast period was partitioned into six non-overlapping eight-hour blocks. The first block was predicted directly from the encoded patient history, static covariates, and the corresponding candidate future action sequence. Each subsequent block was produced by a block-specific decoder conditioned on the encoded history, the future actions assigned to that block, and the preceding predicted block:
\begin{align}
\widehat{\mathbf Y}^{(1)}
&=
f_1\!\left(
\mathbf h,
\mathbf A^{(1)},
\mathbf s
\right),\\
\widehat{\mathbf Y}^{(k)}
&=
f_k\!\left(
\mathbf h,
\widehat{\mathbf Y}^{(k-1)},
\mathbf A^{(k)},
\mathbf s
\right),
\qquad k=2,\ldots,6,
\end{align}
where
$\widehat{\mathbf Y}^{(k)}
=
\widehat{\mathbf Y}_{8(k-1)+1:8k}$
and
$\mathbf A^{(k)}
=
\mathbf A_{8(k-1)+1:8k}$.
Here, $\mathbf h$ is the inherited TFT-action history encoding, $\mathbf A$ is the supplied candidate future action path, and $\mathbf s$ contains static predictive covariates.

Unlike \method{}, the comparator does not use the TFT future forecast as a direct prediction anchor. It also does not use bounded residual correction, residual budgets, state distillation, a discrepancy bottleneck, a local hemodynamic branch, a shared recursive correction decoder, or prediction-level fusion. It therefore isolates a blockwise direct--recursive alternative rather than reproducing \method{}'s constrained correction mechanism. The implementation is literature inspired and is not an exact reproduction of Stratify or ProNet.

\paragraph{Strengthened training and checkpoint selection.}
To avoid disadvantaging the comparator through insufficient optimization, we used a two-stage training protocol with up to 80 epochs and validation-based early stopping. All runs used seeds 42, 43, and 44. The first five epochs trained only the six block decoders. Beginning at epoch 6, the inherited TFT history encoding path was unfrozen and fine-tuned at one tenth of the decoder learning rate. The decoder and inherited-encoder learning rates were $10^{-3}$ and $10^{-4}$, respectively.

Training could stop only after at least 20 epochs. Early stopping used patience 12 and required a minimum validation improvement of $10^{-5}$. Every epoch checkpoint was retained. Validation inference used 70,000 factual windows and 20,000 action-audit windows, but only factual validation metrics entered checkpoint selection. No TEST output was accessed during training or validation selection.

Each seed produced one training trajectory rather than three independently retrained models. Endpoint-heavy, equal, and MAP-heavy criteria selected different saved epochs from that common trajectory. The criteria used the same within-run min--max combinations defined in Section~\ref{app:checkpoint-robustness}: endpoint-heavy weighted normalized endpoint MSE and MAP MAE by $0.80/0.20$, equal weighting used $0.50/0.50$, and MAP-heavy weighting used $0.20/0.80$. These criteria were applied to saved validation checkpoints before TEST evaluation.

Before TEST inference, the selected epoch, checkpoint identity and SHA256, architecture configuration, and evaluator and worker identities were frozen. The MIMIC-IV test roster, feature order, reporting horizons, donor mappings, changed-window masks, and metric denominators were identical to those used in the shared checkpoint-selection analysis. Across \method{}, TFT-action, and DirRecMO-8, all 27 criterion--model--seed evaluation payloads passed exact identity, roster, donor, changed-mask, horizon, and denominator checks.

\begin{table*}[!t]
\centering
\small
\setlength{\tabcolsep}{5pt}
\begin{tabular}{lcccc}
\toprule
Seed &
Epochs run &
Endpoint-heavy epoch &
Equal epoch &
MAP-heavy epoch \\
\midrule
42 & 73 & 61 & 27 & 16 \\
43 & 55 & 54 & 43 & 20 \\
44 & 69 & 57 & 37 & 16 \\
\bottomrule
\end{tabular}
\caption{Convergence and checkpoint selection for the strengthened action-conditioned DirRecMO-8 comparator. All runs used a five-epoch decoder-only warm-up followed by low-learning-rate fine-tuning of the inherited TFT history encoder. Training stopped after 12 validation evaluations without sufficient improvement.}
\label{tab:dirrecmo-convergence}
\end{table*}

All three runs terminated through validation-based early stopping, and every criterion-specific checkpoint was selected from an interior epoch rather than the final training boundary. The endpoint-heavy criterion generally selected later epochs, whereas the MAP-heavy criterion selected earlier epochs with stronger emphasis on MAP validation error.

\paragraph{Primary factual comparison.}
Primary comparisons report \method{} minus DirRecMO-8. Negative values therefore indicate lower error for \method{}. Inference used a stay-clustered, seed-aware hierarchical bootstrap with 2,000 replicates. Benjamini--Hochberg correction was applied across the 18 primary factual comparisons formed by two metrics, three criteria, and three horizons.

\begin{table*}[!t]
\centering
\scriptsize
\setlength{\tabcolsep}{2.8pt}
\resizebox{\textwidth}{!}{%
\begin{tabular}{llccc}
\toprule
Checkpoint criterion &
Metric &
8 h difference [95\% CI] &
24 h difference [95\% CI] &
48 h difference [95\% CI] \\
\midrule
Endpoint-heavy
& Endpoint MSE
& $-0.6998$ [$-1.5101,-0.0801$]
& $-0.8794$ [$-1.9279,-0.0611$]
& $-0.0587$ [$-0.0917,-0.0318$] \\

Endpoint-heavy
& MAP MAE
& $-0.0982$ [$-0.1294,-0.0675$]
& $-0.2510$ [$-0.2882,-0.2179$]
& $-0.3050$ [$-0.3681,-0.2427$] \\

\addlinespace[2pt]

Equal
& Endpoint MSE
& $-0.7404$ [$-1.5880,-0.0914$]
& $-0.9555$ [$-2.1222,-0.0705$]
& $-0.0638$ [$-0.1025,-0.0340$] \\

Equal
& MAP MAE
& $-0.0645$ [$-0.0958,-0.0345$]
& $-0.1563$ [$-0.1929,-0.1232$]
& $-0.1751$ [$-0.2180,-0.1353$] \\

\addlinespace[2pt]

MAP-heavy
& Endpoint MSE
& $-0.7783$ [$-1.6466,-0.1112$]
& $-1.0385$ [$-2.3195,-0.0814$]
& $-0.0813$ [$-0.1383,-0.0398$] \\

MAP-heavy
& MAP MAE
& $-0.0669$ [$-0.1011,-0.0375$]
& $-0.0834$ [$-0.1209,-0.0496$]
& $-0.0838$ [$-0.1164,-0.0527$] \\
\bottomrule
\end{tabular}%
}
\caption{Primary factual comparison of \method{} with the strengthened action-conditioned DirRecMO-8 comparator under three shared checkpoint-selection criteria. Entries are \method{} minus DirRecMO-8; negative values favor \method{}. Endpoint MSE is normalized and unitless, whereas MAP MAE differences are in mmHg. Brackets report stay-clustered, seed-aware hierarchical-bootstrap 95\% confidence intervals. All 18 comparisons had $q=0.001161$ after Benjamini--Hochberg correction within the primary factual comparison family.}
\label{tab:dirrecmo-primary}
\end{table*}

Strengthening DirRecMO-8 through longer training, validation-based early stopping, and low-learning-rate fine-tuning of its inherited history encoder did not eliminate \method{}'s factual advantage. Across endpoint-heavy, equal, and MAP-heavy checkpoint selection, \method{} had lower endpoint multivariate MSE and lower MAP MAE at 8, 24, and 48 hours. All 18 confidence intervals excluded zero after correction.

As a convergence sanity check, the strengthened DirRecMO-8 comparator approximately matched TFT-action in eight-hour MAP accuracy under all three criteria. Its DirRecMO-8-minus-TFT-action differences ranged from $-0.0311$ to $-0.0055$ mmHg at eight hours, and all corresponding intervals included zero. At 24 and 48 hours, its MAP error generally increased relative to TFT-action as recursive block propagation extended across additional forecast blocks. It therefore approximately matched the direct comparator over the first block but did not retain that accuracy at longer horizons.

\paragraph{Action-changing-window factual accuracy.}
We next restricted evaluation to windows in which donor substitution changed at least one supplied future exposure. Factual differences compare observed-target MAP MAE under the recorded future action path and are reported as \method{} minus DirRecMO-8. Negative values favor \method{}. Benjamini--Hochberg correction was applied within the nine-comparison changed-window factual family.

\begin{table*}[!t]
\centering
\scriptsize
\setlength{\tabcolsep}{3.0pt}
\resizebox{\textwidth}{!}{%
\begin{tabular}{lccc}
\toprule
Checkpoint criterion &
8 h difference [95\% CI]; $q$ &
24 h difference [95\% CI]; $q$ &
48 h difference [95\% CI]; $q$ \\
\midrule
Endpoint-heavy
& $-0.2821$ [$-0.4190,-0.1740$]; $q=0.0014$
& $-0.3120$ [$-0.3928,-0.2447$]; $q=0.0014$
& $-0.3095$ [$-0.3848,-0.2325$]; $q=0.0014$ \\

Equal
& $-0.2656$ [$-0.3939,-0.1541$]; $q=0.0014$
& $-0.2104$ [$-0.2943,-0.1389$]; $q=0.0014$
& $-0.1602$ [$-0.2235,-0.0957$]; $q=0.0014$ \\

MAP-heavy
& $-0.2563$ [$-0.3968,-0.1327$]; $q=0.0014$
& $-0.1350$ [$-0.2299,-0.0604$]; $q=0.0014$
& $-0.0749$ [$-0.1198,-0.0294$]; $q=0.0036$ \\
\bottomrule
\end{tabular}%
}
\caption{Factual observed-target MAP comparison in windows where donor substitution changed the supplied future action path. Entries are \method{} minus the strengthened action-conditioned DirRecMO-8 comparator in mmHg; negative values favor \method{}. Brackets report stay-clustered, seed-aware hierarchical-bootstrap 95\% confidence intervals. Reported $q$ values use Benjamini--Hochberg correction within the nine-comparison changed-window factual family.}
\label{tab:dirrecmo-changed-factual}
\end{table*}

The changed-window factual results preserve the primary conclusion. \method{} has lower recorded-path observed-target MAP error at every horizon under all three checkpoint-selection criteria. All nine confidence intervals exclude zero, and all adjusted $q$ values remain below 0.01.

\paragraph{Action-changing-window difference-in-gap.}
For each model, the changed-window action gap is shifted-path observed-target MAP error minus recorded-path observed-target MAP error. The comparison below is the \method{} gap minus the DirRecMO-8 gap. Positive values indicate a larger gap for \method{}, whereas negative values indicate a larger gap for DirRecMO-8. Benjamini--Hochberg correction was applied within this nine-comparison family.

\begin{table*}[!t]
\centering
\scriptsize
\setlength{\tabcolsep}{3.0pt}
\resizebox{\textwidth}{!}{%
\begin{tabular}{lccc}
\toprule
Checkpoint criterion &
8 h difference [95\% CI]; $q$ &
24 h difference [95\% CI]; $q$ &
48 h difference [95\% CI]; $q$ \\
\midrule
Endpoint-heavy
& $+0.2162$ [$+0.0389,+0.3427$]; $q=0.0015$
& $-0.1185$ [$-0.2026,-0.0389$]; $q=0.0015$
& $-0.2613$ [$-0.3686,-0.1325$]; $q=0.0015$ \\

Equal
& $+0.2628$ [$+0.1209,+0.3629$]; $q=0.0015$
& $-0.0391$ [$-0.1490,+0.0539$]; $q=0.5169$
& $-0.1297$ [$-0.2569,+0.0243$]; $q=0.1208$ \\

MAP-heavy
& $+0.2119$ [$+0.1039,+0.2889$]; $q=0.0015$
& $+0.0183$ [$-0.1728,+0.1642$]; $q=0.7366$
& $-0.1482$ [$-0.2141,-0.0738$]; $q=0.0015$ \\
\bottomrule
\end{tabular}%
}
\caption{Difference-in-gap comparison between \method{} and the strengthened action-conditioned DirRecMO-8 comparator in action-changing windows. For each model, the gap is shifted-path observed-target MAP error minus recorded-path observed-target MAP error. Entries report the \method{} gap minus the DirRecMO-8 gap in mmHg. Positive values indicate a larger gap for \method{}; negative values indicate a larger gap for DirRecMO-8. Brackets report stay-clustered, seed-aware hierarchical-bootstrap 95\% confidence intervals, and $q$ values use Benjamini--Hochberg correction within this nine-comparison family.}
\label{tab:dirrecmo-gap}
\end{table*}

\method{} has a larger changed-window action gap at eight hours under all three checkpoint-selection criteria. At 24 and 48 hours, however, the comparison is mixed, and DirRecMO-8 sometimes exhibits a larger gap despite substantially worse factual accuracy. This pattern shows that a larger response to action-path replacement does not by itself imply more accurate or better-calibrated use of the supplied action path. Unconstrained blockwise recursive propagation can amplify action perturbations together with accumulated forecasting error.

\paragraph{Parameters and computational cost.}
The strengthened DirRecMO-8 comparator contains 1,541,263 parameters: 1,328,767 parameters in the inherited TFT history model and 212,496 newly added parameters in the six block-specific decoders. The block size is eight hours and the maximum block count is six. The model does not use the TFT future prediction as an anchor. Decoder-only warm-up lasts five epochs, after which the inherited encoder is fine-tuned. Training times were 3,476.41, 2,597.64, and 3,285.03 seconds for seeds 42, 43, and 44, respectively. These timings describe the strengthened comparator training runs and were not used for model or checkpoint selection.

\FloatBarrier
\section{Action-Path Replacement Analyses}
\label{app:replacement-analyses}

\subsection{Matched Patient-Shift Validity}
\label{app:shift-validity}

The same locked eligibility mask is used for full-stream and future-only protocols. One stay is ineligible for exact 8-hour derangement within its mask stratum. It is excluded only from matched-shift statistics and remains in observed forecasting. Across models, observed sets are identical, shifted sets are identical, and shifted sets equal the locked eligibility mask.

\begin{table}[tb]
\centering
\small
\begin{tabular}{lr}
\toprule
Audit quantity & Result\\
\midrule
Exact time-mask match & 100\%\\
Exact transition-mask match & 100\%\\
Included self-donors & 0\\
Non-derangeable stays & 1/6,046\\
Models with identical shifted sets & All\\
Full/future-only eligible sets equal & Yes\\
Outcome or prediction used in matching & No\\
Shift replicates & 5\\
\bottomrule
\end{tabular}
\caption{Validity summary for the locked matched patient-action control.}
\label{tab:shift-audit}
\end{table}

For DRIFT, TFT-action, and TFT-Large, future-only rows from the frozen action analysis were revalidated against the unified all-valid outputs by exact equality across 46,878 observed rows per model; the maximum observed-metric difference was zero. \tftobj{} future-only outputs were generated within the same unified analysis. Because binary exposure is sparse, a separate realized-change audit quantifies how often the complete stream and horizon-specific future path differ after derangement.

\subsubsection{Absolute Action-Gap Results}

Table~\ref{tab:action-gap} reports the absolute change in factual error after patient-action replacement. These values measure predictive dependence on the supplied path and should be interpreted together with the realized-change rates and the paired difference-in-gap analysis.

\begin{table*}[!t]
\centering
\scriptsize
\setlength{\tabcolsep}{3.6pt}
\begin{tabular}{llcc}
\toprule
Protocol & Model & Endpoint-MSE gap [95\% CI] & MAP-MAE gap (mmHg) [95\% CI]\\
\midrule
Full stream & TFT-action & 0.01114 [0.01026, 0.01220] & 0.0973 [0.0860, 0.1086]\\
Full stream & \tftlarge{} & 0.01064 [0.00980, 0.01140] & 0.0925 [0.0850, 0.1001]\\
Full stream & \tftobj{} & 0.01177 [0.01107, 0.01251] & 0.1465 [0.1331, 0.1632]\\
Full stream & \method{} & 0.01148 [0.01054, 0.01256] & 0.1237 [0.1144, 0.1334]\\
\midrule
Future only & TFT-action & 0.01265 [0.01184, 0.01353] & 0.1063 [0.0982, 0.1147]\\
Future only & \tftlarge{} & 0.01189 [0.01116, 0.01265] & 0.0996 [0.0924, 0.1067]\\
Future only & \tftobj{} & 0.01266 [0.01191, 0.01338] & 0.1556 [0.1399, 0.1752]\\
Future only & \method{} & 0.01296 [0.01217, 0.01383] & 0.1318 [0.1211, 0.1435]\\
\bottomrule
\end{tabular}
\caption{MIMIC-IV all-valid absolute action gaps, defined as shifted minus recorded error and aggregated over seeds and horizons. A larger gap indicates stronger predictive dependence, not better forecasting or causal validity.}
\label{tab:action-gap}
\end{table*}

\subsection{All-Valid Difference-in-Gap Details}
\label{app:gap-details}

\begin{table*}[!t]
\centering
\scriptsize
\setlength{\tabcolsep}{3.2pt}
\begin{tabular}{llrr}
\toprule
Protocol & Comparison / metric & Difference in gap [95\% CI] & $p_{\mathrm{boot}}$\\
\midrule
Full stream & DRIFT vs TFT-action / endpoint MSE & 0.000340 [0.000256, 0.000443] & $\leq0.001$\\
Full stream & DRIFT vs TFT-action / MAP & 0.02635 [0.01577, 0.04081] & $\leq0.001$\\
Full stream & DRIFT vs TFT-Large / endpoint MSE & 0.000844 [$-0.000247$, 0.001814] & 0.109\\
Full stream & DRIFT vs TFT-Large / MAP & 0.03119 [0.02251, 0.04119] & $\leq0.001$\\
Full stream & DRIFT vs TFT-Large-FL / endpoint MSE & $-0.000283$ [$-0.001195$, 0.000781] & 0.580\\
Full stream & DRIFT vs TFT-Large-FL / MAP & $-0.02288$ [$-0.03272,-0.01437$] & $\leq0.001$\\
\midrule
Future only & DRIFT vs TFT-action / endpoint MSE & 0.000316 [0.000210, 0.000415] & $\leq0.001$\\
Future only & DRIFT vs TFT-action / MAP & 0.02545 [0.01626, 0.03872] & $\leq0.001$\\
Future only & DRIFT vs TFT-Large / endpoint MSE & 0.001074 [0.000179, 0.001882] & 0.011\\
Future only & DRIFT vs TFT-Large / MAP & 0.03218 [0.02308, 0.04369] & $\leq0.001$\\
Future only & DRIFT vs TFT-Large-FL / endpoint MSE & 0.000299 [$-0.000314$, 0.001010] & 0.348\\
Future only & DRIFT vs TFT-Large-FL / MAP & $-0.02382$ [$-0.03478,-0.01501$] & $\leq0.001$\\
\bottomrule
\end{tabular}
\caption{All-valid four-arm difference-in-gap inference. Positive values mean DRIFT is more action dependent; negative values mean the comparator has a larger gap.}
\label{tab:gap-paired}
\end{table*}

\subsection{Horizon-Specific Difference-in-Gap}
\label{app:horizon-gap}

\begin{table*}[!t]
\centering
\scriptsize
\setlength{\tabcolsep}{2.8pt}
\begin{tabular}{lllrrr}
\toprule
Protocol & Comparison & Metric & 8h difference [95\% CI] & 24h difference [95\% CI] & 48h difference [95\% CI]\\
\midrule
Full stream & DRIFT vs TFT-action & Endpoint MSE & 0.000165 [0.000112, 0.000228] & 0.000312 [0.000224, 0.000418] & 0.000544 [0.000339, 0.000826]\\
Full stream & DRIFT vs TFT-action & MAP MAE & 0.01954 [0.01527, 0.02456] & 0.02821 [0.01533, 0.04261] & 0.03129 [0.00928, 0.06006]\\
Full stream & DRIFT vs TFT-Large & Endpoint MSE & 0.000050 [$-0.000283$, 0.000379] & 0.000673 [$-0.000263$, 0.001619] & 0.001811 [$-0.000544$, 0.004599]\\
Full stream & DRIFT vs TFT-Large & MAP MAE & 0.02005 [0.01481, 0.02541] & 0.03190 [0.02440, 0.03954] & 0.04162 [0.01400, 0.07127]\\
\midrule
Future only & DRIFT vs TFT-action & Endpoint MSE & 0.000141 [0.000038, 0.000274] & 0.000280 [0.000179, 0.000396] & 0.000527 [0.000310, 0.000823]\\
Future only & DRIFT vs TFT-action & MAP MAE & 0.02038 [0.01758, 0.02337] & 0.02638 [0.01388, 0.03894] & 0.02960 [0.00913, 0.05477]\\
Future only & DRIFT vs TFT-Large & Endpoint MSE & 0.000467 [0.000226, 0.000664] & 0.000787 [0.000122, 0.001555] & 0.001969 [$-0.000397$, 0.004364]\\
Future only & DRIFT vs TFT-Large & MAP MAE & 0.02373 [0.01835, 0.02954] & 0.03181 [0.02052, 0.04238] & 0.04101 [0.01405, 0.07058]\\
\bottomrule
\end{tabular}
\caption{Horizon-specific DRIFT-minus-comparator difference-in-gap using common-stay paired bootstrap intervals. Positive values indicate stronger predictive dependence for DRIFT. MAP differences are in mmHg.}
\label{tab:horizon-gap}
\end{table*}

\subsection{Changed-Window Accuracy and Action-Gap Audit}
\label{app:changed-window-audit}

The all-valid replacement analyses average assignments that alter the supplied action path together with no-op assignments, most commonly all-zero to all-zero replacements. We therefore conducted an additional MIMIC-IV analysis restricted to future-only replacement windows in which the donor path actually changes at least one hourly exposure. This analysis uses the frozen \method{} and TFT-action checkpoints for seeds 42, 43, and 44, the same five fixed one-to-one donor mappings, the same test roster, and the same 8-, 24-, and 48-hour reporting horizons as the primary action audit. No model was retrained or selected, and no eICU file was accessed for this analysis.

For recipient stay $i$, forecast window $w$, donor mapping $d$, and horizon $h$, define
\begin{equation}
C_{iwd}^{(h)}
=
\mathbb{1}\!\left[
\exists k\in\{1,\ldots,h\}:
a^{\mathrm{recorded}}_{iw,k}
\neq
a^{\mathrm{shifted}}_{iwd,k}
\right].
\label{eq:changed-window}
\end{equation}
Only windows with $C_{iwd}^{(h)}=1$ enter this audit. Within each donor mapping, endpoint errors are averaged across changed windows for each stay; available donor-specific stay estimates are then averaged with equal donor weight. Inference first aligns the common stay roster across the three model seeds and then uses 5,000 paired hierarchical bootstrap samples that resample both stays and seeds. Fixed name-derived random seeds make the bootstrap independent of table-row order. Plus-one two-sided probabilities are adjusted separately within each three-horizon MAP comparison family using the Benjamini--Hochberg procedure.

The factual comparison uses observed-target MAP MAE under the recorded path. For model $m$, the changed-window action gap is
\begin{equation}
G_{m}^{(h)}
=
\operatorname{MAE}^{(h)}_{m,\mathrm{shifted}}
-
\operatorname{MAE}^{(h)}_{m,\mathrm{recorded}},
\label{eq:changed-window-gap}
\end{equation}
and the comparative estimand is
$G_{\mathrm{DRIFT}}^{(h)}-G_{\mathrm{TFT}}^{(h)}$.
Positive difference-in-gap values indicate greater degradation for \method{} when the recorded path is replaced by the donor-substituted path; they do not establish that the donor path is a valid alternative-treatment trajectory.

\begin{table*}[!t]
\centering
\scriptsize
\setlength{\tabcolsep}{2.2pt}
\resizebox{\textwidth}{!}{%
\begin{tabular}{lrrrrrr}
\toprule
Horizon &
\method{} factual &
TFT-action factual &
Factual difference [95\% CI] &
\method{} gap &
TFT-action gap &
Difference-in-gap [95\% CI]\\
\midrule
8 h
& 9.011
& 9.108
& $-0.096$ [$-0.125,-0.068$]
& 0.554
& 0.300
& $+0.254$ [$+0.222,+0.285$]\\
24 h
& 9.926
& 9.987
& $-0.061$ [$-0.097,-0.026$]
& 0.713
& 0.449
& $+0.264$ [$+0.174,+0.337$]\\
48 h
& 10.668
& 10.706
& $-0.038$ [$-0.076,+0.002$]
& 0.811
& 0.599
& $+0.211$ [$+0.123,+0.302$]\\
\bottomrule
\end{tabular}%
}
\caption{Additional MIMIC-IV future-only analysis restricted to windows in which donor substitution changes at least one supplied future exposure. All quantities are observed-target MAP MAE in mmHg. Factual values use the recorded path. Gaps are shifted minus recorded error. Factual differences and difference-in-gap values are \method{} minus TFT-action. Negative factual differences favor \method{}; positive difference-in-gap values indicate greater degradation under the substituted path. Within-family corrected factual-difference $q$ values are 0.0012, 0.0012, and 0.0616 at 8, 24, and 48 hours; all three difference-in-gap $q$ values are 0.0004.}
\label{tab:changed-window-map}
\end{table*}

Under the recorded path, \method{} has lower observed-target MAP MAE than TFT-action by 0.096 mmHg at 8 hours and 0.061 mmHg at 24 hours; both paired intervals exclude zero, and both comparisons remain significant after correction across the three MAP horizons. The 48-hour difference remains favorable but uncertain. For shifted-minus-recorded error, the additional degradation for \method{} relative to TFT-action is 0.211--0.264 mmHg, and all three paired intervals exclude zero. The changed-window result therefore combines two properties on the same changed-window subset: better factual forecasting at the short and intermediate horizons and greater loss of accuracy under the donor-substituted path. Robustness of these changed-window conclusions to the three shared checkpoint-selection criteria is reported in Table~\ref{tab:checkpoint-changed-window-robustness}.

\subsection{First-Divergence Temporal Alignment}
\label{app:first-divergence}

A larger action gap does not show when the prediction changes. We therefore align recorded-path and substituted-path predictions to the first future hour at which their raw binary exposure paths differ. For every changed window,
\begin{equation}
\tau_{iwd}
=
\min\left\{
k\in\{1,\ldots,h\}:
a^{\mathrm{recorded}}_{iw,k}
\neq
a^{\mathrm{shifted}}_{iwd,k}
\right\},
\label{eq:first-divergence}
\end{equation}
and relative lead $r=k-\tau_{iwd}$ is negative before divergence and zero at the first divergence. The absolute MAP prediction response is
\begin{equation}
R^{\mathrm{MAP}}_{m,iwd}(r)
=
\left|
\widehat y^{\mathrm{shifted}}_{m,iwd,\tau_{iwd}+r}
-
\widehat y^{\mathrm{recorded}}_{m,iw,\tau_{iwd}+r}
\right|.
\label{eq:aligned-map-response}
\end{equation}
As in the changed-window audit, responses are averaged within stay and donor mapping before seed--stay hierarchical inference. The pre-divergence summary averages $r<0$, whereas the at-and-post-divergence summary averages $r\geq0$. Stays without finite support in a requested phase are excluded explicitly rather than through implicit empty-slice averaging.

The six stored model--seed outputs passed exact-equality checks for the test roster, donor mappings, horizons, feature order, changed-window counts, and phase denominators. The corrected statistical pass read these frozen outputs without additional training, checkpoint selection, model inference, GPU use, or eICU access.

\begin{table*}[!t]
\centering
\scriptsize
\setlength{\tabcolsep}{2.1pt}
\resizebox{\textwidth}{!}{%
\begin{tabular}{llrrrrr}
\toprule
Model &
Horizon &
Pre-divergence [95\% CI] &
At/post-divergence [95\% CI] &
Pre/post ratio [95\% CI] &
Post-minus-pre [95\% CI] &
Stays\\
\midrule
\method{}
& 8 h
& 0.0221 [0.0213, 0.0228]
& 2.591 [2.448, 2.712]
& 0.854\% [0.791\%, 0.926\%]
& 2.569 [2.425, 2.691]
& 2,972\\
\method{}
& 24 h
& 0.0181 [0.0175, 0.0186]
& 3.428 [3.260, 3.573]
& 0.528\% [0.493\%, 0.570\%]
& 3.410 [3.242, 3.555]
& 2,972\\
\method{}
& 48 h
& 0.0170 [0.0164, 0.0176]
& 4.019 [3.876, 4.177]
& 0.423\% [0.402\%, 0.451\%]
& 4.002 [3.858, 4.160]
& 2,359\\
\midrule
TFT-action
& 8 h
& 0.0218 [0.0210, 0.0228]
& 2.017 [1.925, 2.101]
& 1.083\% [1.002\%, 1.180\%]
& 1.995 [1.902, 2.080]
& 2,972\\
TFT-action
& 24 h
& 0.0187 [0.0182, 0.0193]
& 2.626 [2.545, 2.703]
& 0.712\% [0.688\%, 0.737\%]
& 2.607 [2.527, 2.684]
& 2,972\\
TFT-action
& 48 h
& 0.0172 [0.0168, 0.0179]
& 3.248 [3.122, 3.403]
& 0.531\% [0.517\%, 0.545\%]
& 3.231 [3.105, 3.386]
& 2,359\\
\bottomrule
\end{tabular}%
}
\caption{First-divergence-aligned absolute MAP prediction changes in mmHg. Pre-divergence means average relative leads $r<0$; at/post-divergence means average $r\geq0$. The ratio is the pre-divergence mean divided by the at/post-divergence mean. All six post-minus-pre comparisons have within-family $q=0.0004$. Both models show negligible anticipatory change relative to their response once the supplied paths diverge.}
\label{tab:no-anticipation-map}
\end{table*}

\begin{table*}[!t]
\centering
\scriptsize
\setlength{\tabcolsep}{3.0pt}
\begin{tabular}{lrrrrr}
\toprule
Horizon &
\method{} at/post response &
TFT-action at/post response &
Difference [95\% CI] &
Relative to TFT-action &
Stays\\
\midrule
8 h
& 2.591
& 2.017
& $+0.574$ [$+0.521,+0.621$]
& $+28.5\%$
& 2,972\\
24 h
& 3.428
& 2.626
& $+0.802$ [$+0.622,+0.980$]
& $+30.5\%$
& 2,972\\
48 h
& 4.019
& 3.248
& $+0.771$ [$+0.508,+0.970$]
& $+23.7\%$
& 2,359\\
\bottomrule
\end{tabular}
\caption{Paired comparison of absolute MAP prediction response at and after the first supplied-action divergence. The difference is \method{} minus TFT-action in mmHg. All three within-family corrected $q$ values are 0.0004.}
\label{tab:post-divergence-map}
\end{table*}

Both models preserve temporal ordering: prediction differences before the supplied paths diverge are no more than 1.1\% of their at/post-divergence values. However, \method{} has a 23.7--30.5\% larger at/post-divergence MAP response than TFT-action. This larger response is consistent with the changed-window error-gap result in Table~\ref{tab:changed-window-map}: the model changes its MAP forecast more once the supplied action changes, and the mismatched path produces greater degradation relative to recorded-path forecasting.

\begin{table*}[!t]
\centering
\scriptsize
\setlength{\tabcolsep}{2.5pt}
\resizebox{\textwidth}{!}{%
\begin{tabular}{lrrrrr}
\toprule
Horizon &
Divergence hour &
1--2 h after &
3--6 h after &
7--12 h after &
$\geq$13 h after\\
\midrule
8 h
& $-0.083$ [$-0.449,+0.225$]
& $+0.568$ [$+0.371,+0.689$]
& $+0.704$ [$+0.631,+0.792$]
& $+0.722$ [$+0.569,+0.827$]
& \na\\
24 h
& $+0.305$ [$+0.101,+0.659$]
& $+0.994$ [$+0.655,+1.231$]
& $+1.171$ [$+0.879,+1.451$]
& $+0.985$ [$+0.818,+1.248$]
& $+0.486$ [$+0.389,+0.586$]\\
48 h
& $+0.537$ [$+0.264,+0.918$]
& $+1.224$ [$+0.773,+1.532$]
& $+1.385$ [$+0.907,+1.729$]
& $+1.199$ [$+0.871,+1.473$]
& $+0.532$ [$+0.367,+0.749$]\\
\bottomrule
\end{tabular}%
}
\caption{Phase-resolved \method{}-minus-TFT-action difference in absolute MAP prediction response, in mmHg with 95\% paired bootstrap intervals. Relative lead zero is the first hour at which the supplied raw action paths differ. All comparisons except the 8-hour divergence-hour cell remain significant after correction within the phase--horizon MAP family ($q\leq0.000425$); the 8-hour divergence-hour comparison has $q=0.652$. Phase bins are truncated by available horizon support, so the 8-hour 7--12-hour cell contains only relative lead $+7$.}
\label{tab:phase-resolved-map}
\end{table*}

The additional response is largest during the first several hours after divergence and remains positive at longer relative leads. These quantities describe the timing of model prediction changes under supplied-path substitution; they are not estimates of vasopressor onset, duration, or causal treatment effect.

\subsection{Overlapping Stress-Test Action Analysis}
\label{app:action-stress}

The earlier five-group analysis remains a secondary consistency analysis for DRIFT, TFT-action, and TFT-Large. Across 45 overlapping group--horizon--seed cells, full-stream endpoint-MSE advantages are directionally favorable in 45/45 cells for all three models. DRIFT exceeds TFT-action by 0.186 percentage points in normalized endpoint-MSE advantage and 0.246 points in MAP gap; future-only differences are 0.182 and 0.233 points. Relative to TFT-Large, future-only differences are 0.296 and 0.313 points. Because groups overlap and the forecast-loss-transfer control is evaluated in the new all-valid protocol, these counts are not used as the primary comparative inference.

\FloatBarrier
\section{Ablations and Component Diagnostics}
\label{app:component-diagnostics}

\subsection{Complete Seven-Component Ablation Summary}
\label{app:ablation-full}

\begin{table*}[!t]
\centering
\small
\setlength{\tabcolsep}{3.0pt}
\begin{tabular}{lrrrr}
\toprule
Removed component & Endpoint-MSE increase & MAP-MAE increase & Endpoint-gap decrease & MAP-gap decrease\\
\midrule
Recursive path & $+0.022\%$ & $+0.129\%$ & $+0.054\pp$ & $+0.493\pp$\\
State distillation & $+0.013\%$ & $+0.148\%$ & $+0.002\pp$ & $-0.013\pp$\\
Recursive action timing & $+0.001\%$ & $+0.001\%$ & $+0.005\pp$ & $+0.022\pp$\\
Recursive response & $+0.007\%$ & $+0.028\%$ & $+0.036\pp$ & $+0.312\pp$\\
Hemodynamic pathway & $+0.012\%$ & $+0.182\%$ & $+0.011\pp$ & $-0.126\pp$\\
State adapter & $+0.008\%$ & $+0.005\%$ & $+0.006\pp$ & $+0.033\pp$\\
Bottleneck discrepancy correction & $+0.020\%$ & $+0.064\%$ & $+0.020\pp$ & $+0.019\pp$\\
\bottomrule
\end{tabular}
\caption{Complete retrained final-architecture ablation effect sizes. Positive error increase means removal worsens forecasting; positive gap decrease means removal weakens full-stream action-path dependence.}
\label{tab:ablation-full}
\end{table*}

\subsection{Ablation Operator Definitions}
\label{app:ablation-operators}

\paragraph{noRecursivePath.} The transition, recursive action-response network, discrepancy operator, and hemodynamic action projection are removed. The direct anchor, bounded state adapter, and future-action-independent hemodynamic base remain.

\paragraph{noDistillation.} The architecture is unchanged, but $\lambda_{\mathrm{distill}}=0$ in every stage.

\paragraph{noRecursiveActionTiming.} The recursive pathway receives $[a_k,0,0,0]$ instead of exposure, onset, duration, and time-since. Historical and direct-anchor action groups are unchanged.

\paragraph{noRecursiveResponse.} The recursive background transition remains, but the explicit response network, response gate, and hemodynamic action projection are removed.

\paragraph{noHemo.} The mask-decay local hemodynamic base and recursive hemodynamic action correction are removed; the raw-scale hemodynamic objective remains.

\paragraph{noStateAdapter.} The bounded state adapter with a 32-dimensional bottleneck is replaced by zero.

\paragraph{noDiscrepancyCorrection.} Direct and recursive trajectories and distillation remain, but the bottleneck correction added to the direct hidden state is zero.

\subsection{Unified Mask-Decay GRU Re-Inference}
\label{app:grud-unified}

Final unified re-inference of the custom mask-decay GRU gives endpoint MSE of $1.6556\pm0.1160$, $2.5722\pm0.8519$, and $3.0011\pm0.7588$ at 8, 24, and 48 hours; MAP MAE is $9.806\pm0.527$, $15.091\pm4.204$, and $17.324\pm4.258$ mmHg. Target-observed MAP values are $9.889\pm0.551$, $15.292\pm4.351$, and $17.638\pm4.397$.

\subsection{Four-Channel Hemodynamic Results}
\label{app:hemo}

\begin{table*}[!t]
\centering
\scriptsize
\setlength{\tabcolsep}{3.2pt}
\begin{tabular}{llrrrr}
\toprule
Model & Horizon & HR MAE (beats/min) & SBP MAE (mmHg) & DBP MAE (mmHg) & MAP MAE (mmHg)\\
\midrule
DRIFT & 8 & $\best{8.706\pm0.022}$ & $\best{12.078\pm0.016}$ & $\best{8.710\pm0.021}$ & $\best{8.883\pm0.005}$\\
DRIFT & 24 & $\best{10.427\pm0.011}$ & $\best{13.472\pm0.010}$ & $\best{9.690\pm0.013}$ & $\best{9.894\pm0.005}$\\
DRIFT & 48 & $\best{11.856\pm0.006}$ & $\best{14.616\pm0.016}$ & $\best{10.604\pm0.019}$ & $\best{10.613\pm0.015}$\\
\midrule
DRIFT-noHemo & 8 & $\second{8.711\pm0.025}$ & $\second{12.091\pm0.012}$ & $\second{8.817\pm0.029}$ & $\second{8.924\pm0.014}$\\
DRIFT-noHemo & 24 & $\second{10.434\pm0.017}$ & $\second{13.476\pm0.013}$ & $\second{9.764\pm0.027}$ & $\second{9.910\pm0.010}$\\
DRIFT-noHemo & 48 & $\second{11.863\pm0.009}$ & $\second{14.618\pm0.023}$ & $\second{10.645\pm0.021}$ & $\second{10.617\pm0.017}$\\
\midrule
TFT-action & 8 & $8.789\pm0.029$ & $12.127\pm0.015$ & $8.992\pm0.057$ & $8.995\pm0.027$\\
TFT-action & 24 & $10.509\pm0.031$ & $13.510\pm0.006$ & $9.873\pm0.055$ & $9.966\pm0.023$\\
TFT-action & 48 & $11.913\pm0.034$ & $14.636\pm0.020$ & $10.738\pm0.054$ & $10.660\pm0.020$\\
\midrule
TFT-Large & 8 & $8.808\pm0.033$ & $12.132\pm0.003$ & $8.978\pm0.022$ & $8.997\pm0.014$\\
TFT-Large & 24 & $10.528\pm0.038$ & $13.519\pm0.007$ & $9.860\pm0.016$ & $9.975\pm0.014$\\
TFT-Large & 48 & $11.930\pm0.065$ & $14.629\pm0.041$ & $10.711\pm0.020$ & $10.652\pm0.029$\\
\bottomrule
\end{tabular}
\caption{Raw-scale hemodynamic MAE, mean $\pm$ seed SD. Boldface and underlining mark the best and second-best values within each horizon across the four displayed models.}
\label{tab:hemo-absolute}
\end{table*}

\section{Additional Forecasting Analyses and Cost}
\label{app:additional-forecasting}

\subsection{Common-Stay Bootstrap Sensitivity}
\label{app:common-bootstrap}

\begin{table*}[!t]
\centering
\scriptsize
\setlength{\tabcolsep}{2.8pt}
\begin{tabular}{lllrr}
\toprule
Evaluation & Comparison & Metric & Formal difference [95\% CI] & Strict common-cohort difference [95\% CI]\\
\midrule
Standard & DRIFT vs TFT-action & Endpoint MSE & $-0.001215$ [$-0.002718,-0.000212$] & $-0.001209$ [$-0.002632,-0.000291$]\\
Standard & DRIFT vs TFT-action & MAP MAE & $-0.0667$ [$-0.0828,-0.0478$] & $-0.0789$ [$-0.1033,-0.0551$]\\
Standard & DRIFT vs TFT-Large & Endpoint MSE & $-0.001359$ [$-0.003078,0.000510$] & $-0.001115$ [$-0.003898,0.002096$]\\
Standard & DRIFT vs TFT-Large & MAP MAE & $-0.0634$ [$-0.0801,-0.0483$] & $-0.0805$ [$-0.1125,-0.0508$]\\
Observed & DRIFT vs TFT-action & Endpoint MSE & $+0.000312$ [$-0.000371,0.000811$] & $+0.000233$ [$-0.000461,0.000783$]\\
Observed & DRIFT vs TFT-action & MAP MAE & $-0.0512$ [$-0.0688,-0.0314$] & $-0.0612$ [$-0.0862,-0.0384$]\\
Observed & DRIFT vs TFT-Large & Endpoint MSE & $-0.000038$ [$-0.000902,0.000782$] & $-0.000201$ [$-0.001727,0.001154$]\\
Observed & DRIFT vs TFT-Large & MAP MAE & $-0.0470$ [$-0.0637,-0.0334$] & $-0.0639$ [$-0.0949,-0.0347$]\\
\bottomrule
\end{tabular}
\caption{Sensitivity to common-stay resampling. Formal results independently resample paired stays within each seed--horizon cell. The strict sensitivity applies one common stay resample to the all-horizon common roster across sampled seeds. MAP differences are in mmHg; negative values favor DRIFT. Observed common rosters are 3,427 stays for multivariate MSE and 3,346 for MAP.}
\label{tab:common-bootstrap}
\end{table*}

All signs and inferential conclusions are unchanged. Sharing a common stay resample across seeds within each horizon produces the same qualitative result, and restricting to the strict all-horizon common roster preserves the TFT-action endpoint and MAP advantages, the TFT-Large MAP advantage, and the absence of a clear broad observed-entry multivariate difference.

\subsection{Integrated Trajectory Paired Intervals}
\label{app:integrated-ci}

\begin{table*}[!t]
\centering
\small
\setlength{\tabcolsep}{4.2pt}
\resizebox{\textwidth}{!}{%
\begin{tabular}{lrrrr}
\toprule
Model & Integrated trajectory MSE & Integrated MAP MAE & Integrated observed MSE & Integrated observed MAP MAE\\
\midrule
Mask-decay GRU-action & $2.5991\pm0.6154$ & $14.356\pm3.138$ & $2.0589\pm0.2199$ & $14.568\pm3.229$\\
TFT-action & $0.9698\pm0.0002$ & $\second{9.967\pm0.021}$ & $\best{1.2958\pm0.0003}$ & $\second{10.015\pm0.021}$\\
\tftlarge{} & $\second{0.9696\pm0.0014}$ & $9.974\pm0.023$ & $\second{1.2962\pm0.0009}$ & $10.023\pm0.026$\\
\tftobj{} & $0.9943\pm0.0029$ & $10.001\pm0.048$ & $1.3118\pm0.0009$ & $10.047\pm0.046$\\
\textbf{\method{}} & $\best{0.9688\pm0.0012}$ & $\best{9.889\pm0.009}$ & $1.2965\pm0.0002$ & $\best{9.944\pm0.007}$\\
\bottomrule
\end{tabular}
}%
\caption{MIMIC-IV arithmetic means over lead hours 1--48 on the same 3,535 stays, mean $\pm$ seed SD. MAP quantities are in mmHg. Boldface and underlining mark the lowest and second-lowest values.}
\label{tab:integrated-trajectory}
\end{table*}

\begin{table*}[!t]
\centering
\scriptsize
\setlength{\tabcolsep}{2.5pt}
\resizebox{\textwidth}{!}{%
\begin{tabular}{lrrrr}
\toprule
Comparison & Integrated 1--48-h MSE & Integrated MAP MAE (mmHg) & Integrated observed MSE & Integrated observed MAP MAE (mmHg)\\
\midrule
DRIFT vs TFT-action & $-0.000959$ [$-0.002112,-0.000200$] & $-0.0776$ [$-0.1011,-0.0536$] & $+0.000662$ [$0.000227,0.001139$] & $-0.0713$ [$-0.0957,-0.0462$]\\
DRIFT vs TFT-Large & $-0.000727$ [$-0.004073,0.003188$] & $-0.0844$ [$-0.1208,-0.0521$] & $+0.000239$ [$-0.001554,0.001879$] & $-0.0794$ [$-0.1134,-0.0448$]\\
DRIFT vs TFT-Large-FL & $-0.025444$ [$-0.030883,-0.018888$] & $-0.1111$ [$-0.1781,-0.0527$] & $-0.015336$ [$-0.018198,-0.012885$] & $-0.1035$ [$-0.1655,-0.0479$]\\
\bottomrule
\end{tabular}
}%

\caption{Secondary paired common-stay bootstrap differences for integrated 1--48-hour metrics on the fixed 3,535-stay cohort. Each cell is DRIFT minus comparator with a 95\% interval; negative values favor DRIFT.}
\label{tab:integrated-ci}
\end{table*}

\subsection{Common 48-Hour-Eligible Cohort}
\label{app:common-cohort}

\begin{table*}[!t]
\centering
\small
\resizebox{\textwidth}{!}{%
\begin{tabular}{lrrrrrr}
\toprule
& \multicolumn{2}{c}{8 hours} & \multicolumn{2}{c}{24 hours} & \multicolumn{2}{c}{48 hours}\\
\cmidrule(lr){2-3}\cmidrule(lr){4-5}\cmidrule(lr){6-7}
Model & Endpoint MSE & MAP MAE (mmHg) & Endpoint MSE & MAP MAE (mmHg) & Endpoint MSE & MAP MAE (mmHg)\\
\midrule
DRIFT & $\best{1.3089}$ & $\best{8.894}$ & $\best{0.6060}$ & $\best{9.867}$ & $\best{0.7106}$ & $\best{10.613}$\\
TFT-action & 1.3104 & $\second{9.017}$ & $\second{0.6072}$ & $\second{9.948}$ & $\second{0.7114}$ & 10.660\\
TFT-Large & $\second{1.3090}$ & 9.022 & 0.6075 & 9.959 & 0.7131 & $\second{10.652}$\\
\bottomrule
\end{tabular}
}%

\caption{Absolute endpoint results on the same 3,535 stays at all horizons. Boldface and underlining mark the best and second-best values among the three displayed models.}
\label{tab:common-cohort}
\end{table*}

\subsection{Lead-Time and Direct--Recursive Diagnostics}
\label{app:trajectory-details}

The fixed-cohort trajectory files contain 48 lead hours, five models, and three seeds (720 lead-model-seed rows). The integrated values are reported in Table~\ref{tab:integrated-trajectory}. The following table summarizes how strongly the recursive path changes the direct solution.

\begin{table}[tb]
\centering
\small
\resizebox{\columnwidth}{!}{%
\begin{tabular}{rrrr}
\toprule
Lead & Final--direct MSE & Recursive--direct latent MSE & Correction RMS\\
\midrule
1 & 0.001327 & 1.4363 & 0.0188\\
8 & 0.000687 & 0.2688 & 0.0117\\
16 & 0.000660 & 0.1563 & 0.0104\\
24 & 0.000662 & 0.1400 & 0.0096\\
32 & 0.000663 & 0.1353 & 0.0088\\
40 & 0.000671 & 0.1374 & 0.0082\\
48 & 0.000686 & 0.1438 & 0.0076\\
\bottomrule
\end{tabular}
}%

\caption{DRIFT pathway diagnostics by selected lead hour, averaged over three seeds on the fixed 3,535-stay cohort.}
\label{tab:lead-drift}
\end{table}

The recursive/direct latent discrepancy declines sharply with lead, while the applied correction remains small in absolute RMS and decreases with lead; the final output stays close to the direct branch. This supports the constrained-correction interpretation rather than an unconstrained second decoder.

\subsection{Utilization Diagnostics}
\label{app:utilization}

\begin{table*}[!t]
\centering
\scriptsize
\setlength{\tabcolsep}{3.6pt}
\begin{tabular}{llrrrrrr}
\toprule
Mode & Horizon & State adapter RMS & Direct--recursive RMS & Latent correction RMS & Effective response RMS & Hemo action RMS & Final--direct RMS\\
\midrule
Observed & 8 & 0.0130 & 0.8532 & 0.0152 & 0.0273 & 0.0052 & 0.0314\\
Observed & 24 & 0.0129 & 0.5894 & 0.0131 & 0.0254 & 0.0044 & 0.0287\\
Observed & 48 & 0.0128 & 0.4933 & 0.0117 & 0.0281 & 0.0045 & 0.0272\\
\midrule
Zero action & 8 & 0.0130 & 0.8562 & 0.0150 & 0 & 0 & 0.0312\\
Zero action & 24 & 0.0129 & 0.5832 & 0.0126 & 0 & 0 & 0.0281\\
Zero action & 48 & 0.0127 & 0.4771 & 0.0111 & 0 & 0 & 0.0262\\
\bottomrule
\end{tabular}
\caption{Mean diagnostic pathway magnitudes. Action-conditioned response and hemodynamic action correction vanish under zero action.}
\label{tab:utilization}
\end{table*}

The response-gate mean is 0.985--0.989. We do not interpret it as a sharply selective event detector; structural exposure gating and the response vector account for zero-action invariance.

\subsection{Exploratory Subgroup Results}
\label{app:subgroups}

Subgroups use age (18--44, 45--64, 65--79, 80+), sex, five harmonized race groups, and nine ICU-type groups with sufficient sample size. No subgroup-specific bootstrap or multiplicity correction is applied.

\begin{table*}[!t]
\centering
\small
\begin{tabular}{lrr}
\toprule
Comparison & DRIFT lower endpoint MSE & DRIFT lower MAP MAE\\
\midrule
DRIFT vs TFT-action & 57/60 & 55/60\\
DRIFT vs TFT-Large & 39/60 & 56/60\\
DRIFT vs TFT-Large-FL & 60/60 & 58/60\\
\bottomrule
\end{tabular}
\caption{Directional counts across 20 subgroup levels and three horizons (60 cells). These are exploratory descriptive counts, not independent or multiplicity-adjusted tests.}
\label{tab:subgroup-direction}
\end{table*}

Against TFT-action, the five MAP reversals occur at 48 hours for Asian, neuro-intermediate, age 45--64, and neuro-stepdown groups, plus a small 24-hour neuro-intermediate reversal. Several of these groups are small (e.g., Asian 48-hour $n=102$, neuro-stepdown 48-hour $n=89$). Against TFT-Large, four MAP cells reverse. These findings motivate targeted replication and should not be interpreted as evidence of subgroup equivalence or fairness.

\subsection{Parameter, Training, and Inference Cost}
\label{app:cost}

\begin{table*}[!t]
\centering
\scriptsize
\setlength{\tabcolsep}{2.5pt}
\resizebox{\textwidth}{!}{%
\begin{tabular}{lrrrrl}
\toprule
Model / phase & Total parameters & Added parameters & Training type & Hours & Note\\
\midrule
TFT-action pretraining & 1,328,767 & \na & End to end & $0.351\pm0.025$ & Seed-matched initialization\\
DRIFT adaptation & 1,551,342 & 222,575 & After TFT initialization & $0.355\pm0.073$ & New modules plus Stage-4 output tuning\\
Full DRIFT pipeline & 1,551,342 & 222,575 & Approx. end to end & 0.706 & Mean phase times summed; descriptive\\
TFT-Large & 1,586,047 & \na & End to end & $0.350\pm0.036$ & Capacity-matched direct model\\
TFT-Large-FL & 1,586,047 & \na & End to end & $0.359\pm0.012$ & Frozen loss transfer; selected epochs 15/12/13\\
\bottomrule
\end{tabular}%
}
\caption{Recorded shared-server training cost on one H100 NVL per run. Added parameters are relative to the TFT-action initialization. Times include optimization and epoch-level validation but exclude diagnostic or action-replacement inference and statistical aggregation. The full-pipeline time is the descriptive sum of the mean anchor-pretraining and adaptation times.}
\label{tab:training-cost}
\end{table*}

\begin{table}[tb]
\centering
\small
\begin{tabular}{lrrr}
\toprule
Model & ms/batch & ms/window & Peak allocated GiB\\
\midrule
DRIFT & 19.82 & 0.0516 & 0.251\\
TFT-action & 6.52 & 0.0170 & 0.238\\
TFT-Large & 9.90 & 0.0258 & 0.486\\
\bottomrule
\end{tabular}
\caption{Diagnostic shared-H100 inference timing at batch size 384. Concurrent workloads prevent interpretation as an isolated-hardware benchmark.}
\label{tab:inference-cost}
\end{table}

\section{Implementation Audit}
\label{app:implementation-audit}

The reproducibility record binds the formal configuration, evaluation code, model identities, seeds, and hardware policy. All registered primary-cohort model, training, and inference artifact jobs passed signature checks, and all 45 model--seed--horizon cells passed exact identifier and window-count validation. Statistical outputs were independently checked against the underlying subject-level exports.

All ablation cells passed the same identifier and window checks and prohibited epoch-zero fallback. Statistical code fails on duplicate keys, unmatched rows outside the eligibility definition, inconsistent window counts, non-finite metrics, missing protocol rows, or invalid denominators. The artifact retains evaluation source, configurations, audits, tables, figures, and subject-level metric exports; checkpoints are bound by identity and hash.

\subsection{Data Governance, Ethics, and Code Availability}
\label{app:governance}

This retrospective secondary analysis used the de-identified, credentialed-access MIMIC-IV and eICU-CRD databases distributed through PhysioNet. Access followed completion of the required training and acceptance of the applicable data-use agreements. Patient-level data and derived artifacts were stored and analyzed only on authorized institutional computing infrastructure; they were not transmitted to external APIs, uploaded to public cloud storage, or shared with unauthorized individuals. The manuscript contains no patient-level timestamps or identifying information.

An implementation package containing the core model, training, checkpoint-selection, evaluation, and action-path auditing utilities has been prepared. A public version will be released under the MIT License upon publication, together with configuration files and non-sensitive synthetic examples. In accordance with the PhysioNet data-use agreements, neither the implementation package nor the public release redistributes source data, identifiers, split rosters, patient-level tensors or predictions, sensitive checkpoints, or other restricted patient-level artifacts. Authorized researchers must independently obtain credentialed access to the source databases.

\paragraph{Use of generative AI tools.}
Generative AI tools were used to assist with language editing, LaTeX drafting, and code and documentation review. No patient-level data or restricted database content were provided to these tools. The authors independently verified the manuscript, formulas, references, code, analyses, and results, executed all reported experiments, and remain fully responsible for the reported work.

\section{Protocol-Locked eICU-CRD Replication}
\label{app:eicu-protocol}

\subsection{Data Provenance, Development Split, and Sealed Test Cohort}
\label{app:eicu-data-provenance}
eICU-CRD v2.0 contains de-identified multi-center critical-care data and is distributed through PhysioNet \citep{pollard2018eicu,pollard2019eicu20}. The eICU replication pipeline was constructed separately from the primary MIMIC artifacts. All 32 downloaded eICU source files matched the official SHA256 inventory. The split key was \texttt{uniquepid}. For each unique patient, the first 64 bits of \texttt{SHA256(20260712|uniquepid)} were mapped to $[0,1)$ and assigned by fixed 70/15/15 thresholds; no hospital, unit, mortality, demographic, or action stratification was used. All hospital and ICU stays for one patient therefore remained in one split. The hospital--ward eligibility interface and all cohort rules were frozen using development data. Final development tensors contain 38,633 training and 8,115 validation stays; no test clinical row was accessed during feature mapping, preprocessing fitting, model development, or checkpoint selection.

The architecture, cohort rules, target mapping, action reconstruction,
validation-selected checkpoints, metrics, donor-construction algorithm and
seeds, bootstrap, and hypothesis families were fixed independently of test
outcomes. The 36 neural pipelines and Persistence were evaluated without test
refitting or checkpoint reselection. Concrete donor mappings were generated
once from the fixed algorithm, exact test-mask strata, and prespecified seeds.

The sealed test list contained 20,941 patients. Table~\ref{tab:eicu-cohort-flow} reports mutually exclusive exclusions. The final cohort of 8,345 stays exactly matches the test tensor roster. The 8-, 24-, and 48-hour evaluator contains 372,969, 239,449, and 86,171 valid windows from 8,345, 8,337, and 4,486 stays, respectively. The fixed 1--48-hour trajectory roster is the same 4,486-stay set at every lead.

\begin{table}[t]
\centering
\small
\begin{tabular}{lr}
\toprule
Patient-level outcome & Count\\
\midrule
Sealed test patients & 20,941 \\
Multiple hospital stays & 2,852 \\
Hospital/unit ordering discordant & 32 \\
Age below 18 or unparsable & 56 \\
ICU length of stay below 24 hours & 4,971 \\
Death at or before hour 6 & 1 \\
Frozen ward/interface ineligible & 3,362 \\
Pre-index vasopressor exposure & 1,322 \\
Final eligible patients/stays & 8,345 \\
\bottomrule
\end{tabular}
\caption{Locked eICU test cohort flow. Exclusion reasons are mutually exclusive at the patient level; the final 8,345 eligible patients contribute the 8,345 retained stays used for evaluation.}
\label{tab:eicu-cohort-flow}
\end{table}

\subsection{Demographics, Clinical Operations, and Site Overlap}
\label{app:eicu-demographics}

Table~\ref{tab:eicu-test-characteristics} reports aggregated demographics and clinical-operational characteristics of the fixed test roster. This descriptive audit does not alter inference, bootstrap, or FDR outputs and exports no patient-level rows. The median number of action-positive hours is zero when calculated over all stays because 90.99\% have no post-index exposure; 752 stays (9.01\%) have any post-index vasoactive exposure and 326 (3.91\%) initiate exposure during hours 6--12.

\begin{table*}[!t]
\centering
\small
\setlength{\tabcolsep}{4.0pt}
\begin{tabular}{lr lr}
\toprule
Characteristic & Value & Characteristic & Value\\
\midrule
Locked test stays & 8,345 & Unique hospitals & 131\\
Age, mean $\pm$ SD (years) & $63.6\pm17.4$ & Unique hospital--ward pairs & 212\\
Age, median (IQR) & 66 (53--77) & Distinct ICU unit types & 8\\
Female & 3,853 (46.17\%) & Hospital mortality & 705 (8.45\%)\\
Unknown sex & 3 (0.04\%) & ICU mortality & 396 (4.75\%)\\
ICU LOS, median (IQR), days & 2.12 (1.52--3.72) & Hospital LOS, median (IQR), days & 5.85 (3.45--9.85)\\
Any post-index vasoactive exposure & 752 (9.01\%) & Exposure during hours 6--12 & 326 (3.91\%)\\
\bottomrule
\end{tabular}
\caption{Aggregated characteristics of the final 8,345-stay eICU test cohort. LOS denotes length of stay. Counts and percentages are descriptive summaries of the fixed cohort and do not alter the inferential analysis.}
\label{tab:eicu-test-characteristics}
\end{table*}

\begin{table*}[!t]
\centering
\small
\setlength{\tabcolsep}{4.0pt}
\begin{tabular}{lrr lrr}
\toprule
Ethnicity group & Stays & Percentage & ICU unit type & Stays & Percentage\\
\midrule
White/Caucasian & 6,612 & 79.23\% & Med--Surg ICU & 4,729 & 56.67\%\\
Black/African American & 788 & 9.44\% & Neuro ICU & 746 & 8.94\%\\
Other/Unknown & 497 & 5.96\% & Cardiac ICU & 682 & 8.17\%\\
Hispanic/Latino & 312 & 3.74\% & MICU & 645 & 7.73\%\\
Asian & 76 & 0.91\% & CCU--CTICU & 587 & 7.03\%\\
Native American & 60 & 0.72\% & SICU & 555 & 6.65\%\\
 &  &  & CSICU & 288 & 3.45\%\\
 &  &  & CTICU & 113 & 1.35\%\\
\bottomrule
\end{tabular}
\caption{Recorded ethnicity and ICU unit-type distributions in the final eICU test cohort. Ethnicity categories are descriptive database fields and are not used to support subgroup-performance or fairness claims.}
\label{tab:eicu-ethnicity-unit-types}
\end{table*}

\begin{table*}[!t]
\centering
\small
\setlength{\tabcolsep}{3.8pt}
\begin{tabular}{lrrrp{5.2cm}}
\toprule
Site-coverage quantity & Train & Validation & TEST & Overlap conclusion\\
\midrule
Unique hospitals & 131 & 130 & 131 & All 131 test hospitals were represented in training and development; 0 were unseen.\\
Unique hospital--ward pairs & 212 & 210 & 212 & All 212 test hospital--ward pairs were represented in training; 0 were unseen.\\
test stays from training-seen hospitals & \na & \na & 8,345 & 8,345/8,345 test stays (100\%) came from hospitals represented in training.\\
test stays from development-seen hospitals & \na & \na & 8,345 & 8,345/8,345 test stays (100\%) came from hospitals represented in development.\\
\bottomrule
\end{tabular}
\caption{Hospital and hospital--ward overlap under the patient-level split. The database is multi-center, but the cross-database replication is not hospital-held-out: every TEST hospital and hospital--ward pair was already represented in training.}
\label{tab:eicu-site-overlap}
\end{table*}

\subsection{Tensorization and Source-Specific Action Reconstruction}
\label{app:eicu-tensorization}
The external tensor follows the same $72\times27$ hourly layout and the same $A_t\rightarrow X_{t+1}$ convention as the primary study. Values are constrained by frozen plausible ranges, aggregated within hour, forward filled only from the past, and supplemented by training-set fallback medians. Means, standard deviations, fallback values, categorical encodings, and the treatment-likelihood static context were fitted on training stays only. The test tensor stores raw and normalized physiology, direct-observation masks, elapsed time, time/transition masks, action and action-memory tensors, and identifiers in separate arrays. It satisfies finite-value, denominator, pre-index-action, patient-disjointness, and train-statistic-reuse checks.

Vasoactive exposure was reconstructed from positive \texttt{infusionDrug} snapshots for dopamine, epinephrine, norepinephrine, phenylephrine, and vasopressin. A positive snapshot was carried for 120 minutes, clipped to the first 72 ICU hours, and converted to hourly binary overlap. This deterministic rule was frozen on training data and copied to validation and test. It is an operational approximation required by the eICU source, not a claim of exact infusion duration or dose equivalence.

\subsection{eICU Feature Mapping and Direct-Observation Coverage}
\label{app:eicu-features}

\begin{table*}[!t]
\centering
\scriptsize
\setlength{\tabcolsep}{2.4pt}
\resizebox{\textwidth}{!}{%
\begin{tabular}{lllll}
\toprule
Target & Frozen eICU source / precedence & Canonical unit & Plausible range & Conversion and hourly aggregation\\
\midrule
DBP & \texttt{periodic systemic; aperiodic noninvasive fallback} & mmHg & 10--220 & hourly median; invasive preferred \\
GCS eye & \texttt{nurseCharting Glasgow / Eyes} & score & 1--4 & hourly median \\
GCS motor & \texttt{nurseCharting Glasgow / Motor} & score & 1--6 & hourly median \\
GCS verbal & \texttt{nurseCharting Glasgow / Verbal} & score & 1--5 & hourly median \\
Heart rate & \texttt{vitalPeriodic.heartRate} & beats/min & 20--260 & hourly median \\
Direct MAP & \texttt{periodic systemic mean; aperiodic noninvasive fallback} & mmHg & 20--250 & no SBP/DBP-derived MAP \\
Respiratory rate & \texttt{vitalPeriodic.respiration} & breaths/min & 1--100 & hourly median \\
SBP & \texttt{periodic systemic; aperiodic noninvasive fallback} & mmHg & 30--320 & hourly median; invasive preferred \\
SpO$_2$ & \texttt{vitalPeriodic.saO2} & \% & 50--100 & hourly median \\
Temperature & \texttt{nurseCharting; periodic fallback} & $^\circ\mathrm{C}$ & 25--45 & 70--120 converted $^\circ\mathrm{F}$ to $^\circ\mathrm{C}$ \\
Albumin & \texttt{labName=albumin} & g/dL & 0.5--6.5 & hourly median \\
Anion gap & \texttt{labName=anion gap} & mEq/L & 0--60 & mmol/L treated 1:1 \\
Bicarbonate & \texttt{labName=bicarbonate} & mmol/L & 3--60 & HCO3/Total CO$_2$ excluded \\
Total bilirubin & \texttt{labName=total bilirubin} & mg/dL & 0.05--80 & hourly median \\
BUN & \texttt{labName=BUN} & mg/dL & 1--250 & hourly median \\
Chloride & \texttt{labName=chloride} & mmol/L & 60--150 & mEq/L treated 1:1 \\
Creatinine & \texttt{labName=creatinine} & mg/dL & 0.1--25 & hourly median \\
Glucose & \texttt{labName=glucose} & mg/dL & 20--1200 & hourly median \\
Hemoglobin & \texttt{labName=Hgb} & g/dL & 3--25 & hourly median \\
INR & \texttt{labName=PT - INR} & ratio & 0.5--20 & hourly median \\
Lactate & \texttt{labName=lactate} & mmol/L & 0.1--40 & mEq/L treated 1:1 \\
Platelet & \texttt{labName=platelets x 1000} & $10^3$/uL & 1--2000 & unit-label normalization \\
Potassium & \texttt{labName=potassium} & mmol/L & 1--12 & mEq/L treated 1:1 \\
PT & \texttt{labName=PT} & seconds & 5--180 & hourly median \\
PTT & \texttt{labName=PTT} & seconds & 8--300 & hourly median \\
Sodium & \texttt{labName=sodium} & mmol/L & 90--190 & mEq/L treated 1:1 \\
WBC & \texttt{labName=WBC x 1000} & $10^3$/uL & 0.1--400 & unit-label normalization \\
\bottomrule
\end{tabular}
}%
\caption{Frozen eICU mapping, units, plausible ranges, and aggregation for all 27 targets. Nursing and laboratory availability uses the final conservative time rule, taking the later clinical-result and revision/entry offsets when both exist. Values outside these ranges are treated as unavailable; source names, unit rules, and ranges were fixed using development data and retained unchanged during evaluation.}
\label{tab:eicu-feature-source-map}
\end{table*}

All 27 features had nonzero direct observation in the locked test cohort. Table~\ref{tab:eicu-feature-coverage} reports the percentage of final stays with at least one direct observation. Coverage is high for routine vital signs and common chemistry/hematology variables but substantially lower for lactate and coagulation studies, motivating the separate observed-target estimand.

\begin{table*}[!t]
\centering
\small
\begin{tabular}{lr lr}
\toprule
Feature & Stays observed (\%) & Feature & Stays observed (\%)\\
\midrule
DBP & 98.97 & GCS eye & 71.38 \\
GCS motor & 71.36 & GCS verbal & 71.32 \\
Heart rate & 98.66 & Direct MAP & 98.97 \\
Respiratory rate & 92.93 & SBP & 98.96 \\
SpO$_2$ & 98.27 & Temperature & 99.36 \\
Albumin & 54.91 & Anion gap & 74.03 \\
Bicarbonate & 90.86 & Total bilirubin & 49.63 \\
BUN & 95.58 & Chloride & 95.69 \\
Creatinine & 95.63 & Glucose & 95.65 \\
Hemoglobin & 94.82 & INR & 42.98 \\
Lactate & 28.28 & Platelet & 94.51 \\
Potassium & 95.87 & PT & 41.22 \\
PTT & 29.31 & Sodium & 95.75 \\
WBC & 94.51 &  &  \\
\bottomrule
\end{tabular}
\caption{Direct-observation coverage in the final 8,345-stay eICU test cohort. This table reports whether a stay contains at least one direct observation, not hourly measurement density.}
\label{tab:eicu-feature-coverage}
\end{table*}

\subsection{eICU Model Matrix and Frozen Evaluation}
\label{app:eicu-freeze}

The eICU benchmark contains 12 neural architectures and three fixed training seeds, for 36 trainable pipelines. Parameter counts are shown in Table~\ref{tab:eicu-model-matrix}. Checkpoints were selected using factual validation forecasts only; action-input perturbations and test outcomes never entered checkpoint selection. The fixed evaluator reproduced each selected checkpoint's 8-, 24-, and 48-hour validation endpoint MSE, MAP MAE, and exact window count. SHA256 manifests bind the checkpoint whitelist, evaluator, configuration, test tensor, donor mappings, and reported results.

\begin{table*}[!t]
\centering
\scriptsize
\setlength{\tabcolsep}{3.0pt}
\begin{tabular}{lrr}
\toprule
Neural model & Parameters & Seeds\\
\midrule
DRIFT & 1,551,342 & 3 \\
TFT-action & 1,328,767 & 3 \\
TFT-Large & 1,586,047 & 3 \\
TFT-Large-FL & 1,586,047 & 3 \\
iTransformer-action & 1,560,672 & 3 \\
PatchTST-action & 1,196,571 & 3 \\
GRU-action & 649,243 & 3 \\
GRU-no-action & 649,243 & 3 \\
Mask-decay GRU-action & 342,353 & 3 \\
Transformer-action & 690,971 & 3 \\
Transformer-no-action & 690,971 & 3 \\
TFT-no-dynamic-action & 1,328,767 & 3 \\
\bottomrule
\end{tabular}
\caption{Frozen eICU neural benchmark. Every architecture uses three fixed seeds. Persistence is deterministic and is omitted.}
\label{tab:eicu-model-matrix}
\end{table*}

The test tensor, donor mappings, evaluator, and evaluation configuration are hash-bound in the reproducibility artifact.

\subsection{Frozen PatchTST and iTransformer Adaptations}
\label{app:eicu-modern-baselines}

Both modern baselines, PatchTST-action \citep{nie2023patchtst} and iTransformer-action \citep{liu2024itransformer}, use the same 48-hour historical window, candidate future action-memory variables, temporal features, static context, feature-weighted trajectory objective, and factual validation checkpoint criterion as the other direct controls. Their architectures and hyperparameters were fixed using development data; no test outcome informed model selection.

\begin{table*}[!t]
\centering
\scriptsize
\setlength{\tabcolsep}{2.7pt}
\resizebox{\textwidth}{!}{%
\begin{tabular}{p{2.0cm}p{5.2cm}p{3.1cm}p{3.2cm}p{2.4cm}}
\toprule
Model & Action-conditioned input construction & Frozen architecture & Frozen optimization & Checkpoint selection\\
\midrule
iTransformer-action &
Each physiological variable is a token formed from its 48-hour value, observation-mask, and elapsed-time histories. Flattened historical action/time features, zero-padded candidate future action/time features, and static context are added as three context tokens. &
$d_{\mathrm{model}}=208$, 4 encoder layers, 8 heads, feed-forward width 416, dropout 0.10; variable-wise output heads predict up to 48 leads. &
8/24/48-hour horizons sampled equally; at most 18 epochs, 1,000 minibatches/epoch, batch 192, evaluation batch 384, AdamW with learning rate $2{\times}10^{-4}$, weight decay $10^{-4}$, gradient clip 0.5, patience 5. &
Fixed factual validation composite used by the direct controls.\\
PatchTST-action &
Per-hour values, masks, elapsed times, historical action memory, and time features are concatenated before patching. Invalid padded patches are masked. The pooled history representation is concatenated with a lead-specific query built from candidate future action, future time, and static context. &
Patch length 8, stride 4, $d_{\mathrm{model}}=192$, 3 encoder layers, 8 heads, feed-forward width 384, dropout 0.10. &
8/24/48-hour horizons sampled equally; at most 18 epochs, 1,000 minibatches/epoch, batch 192, evaluation batch 384, AdamW with learning rate $2{\times}10^{-4}$, weight decay $10^{-4}$, gradient clip 0.5, patience 5. &
Same fixed factual validation composite; secondary modern baseline with no hyperparameter sweep.\\
\bottomrule
\end{tabular}
}%
\caption{Frozen eICU adaptations of PatchTST and iTransformer. The action-conditioned designs expose both historical action memory and the supplied candidate future path rather than applying the original action-agnostic architectures unchanged.}
\label{tab:eicu-modern-baselines}
\end{table*}

\subsection{Complete eICU Horizon-Specific Forecasting}
\label{app:eicu-horizon-full}

\begin{table*}[!t]
\centering
\scriptsize
\setlength{\tabcolsep}{2.2pt}
\resizebox{\textwidth}{!}{%
\begin{tabular}{lrrrrrr}
\toprule
& \multicolumn{2}{c}{8 hours} & \multicolumn{2}{c}{24 hours} & \multicolumn{2}{c}{48 hours}\\
\cmidrule(lr){2-3}\cmidrule(lr){4-5}\cmidrule(lr){6-7}
Model & Endpoint MSE & MAP MAE & Endpoint MSE & MAP MAE & Endpoint MSE & MAP MAE\\
\midrule
Persistence & 0.4246 & 10.867 & 0.7782 & 12.701 & 1.1344 & 14.182\\
GRU-no-action & $\second{0.3460}$ & 9.105 & 0.6007 & 10.531 & 0.8050 & 11.485\\
GRU-action & 0.3468 & 9.097 & 0.5983 & 10.502 & 0.7976 & 11.398\\
Mask-decay GRU-action & 0.3518 & 9.117 & 0.5987 & 10.525 & 0.7973 & 11.435\\
Transformer-no-action & 0.3477 & 9.087 & 0.5971 & 10.481 & 0.8009 & 11.463\\
Transformer-action & 0.3488 & 9.077 & $\second{0.5951}$ & $\second{10.426}$ & $\second{0.7931}$ & 11.340\\
TFT-no-dynamic-action & $\best{0.3454}$ & 9.093 & 0.5977 & 10.481 & 0.8011 & 11.460\\
TFT-action & 0.3479 & 9.087 & 0.5972 & 10.448 & 0.7949 & $\second{11.336}$\\
\tftlarge{} & 0.3470 & 9.085 & 0.5968 & 10.442 & 0.7977 & 11.347\\
\tftobj{} & 0.3727 & $\second{9.038}$ & 0.6108 & 10.469 & 0.8086 & 11.360\\
PatchTST-action & 0.3587 & 9.160 & 0.5980 & 10.495 & 0.7954 & 11.380\\
iTransformer-action & 0.3503 & 9.161 & 0.6022 & 10.553 & 0.8067 & 11.467\\
\textbf{\method{} (ours)} & 0.3466 & $\best{9.035}$ & $\best{0.5947}$ & $\best{10.412}$ & $\best{0.7927}$ & $\best{11.302}$\\
\bottomrule
\end{tabular}
}%
\caption{Complete eICU horizon-specific locked-test results. Neural entries are seed means; seed standard deviations for the shared core models appear in Table~\ref{tab:cross-dataset-horizon} in the main text, and horizon-averaged mean $\pm$ SD values appear in Table~\ref{tab:eicu-full-ranking}. Boldface and underlining mark the best and second-best values across the complete roster. Endpoint MSE is normalized using eICU training statistics.}
\label{tab:eicu-horizon-full}
\end{table*}

\subsection{Complete eICU Locked-Test Ranking}
\label{app:eicu-ranking}

\begin{table*}[!t]
\centering
\scriptsize
\setlength{\tabcolsep}{3.5pt}
\resizebox{\textwidth}{!}{%
\begin{tabular}{lrrrrr}
\toprule
Model & Seeds & Endpoint MSE & MAP MAE & Observed endpoint MSE & Observed MAP MAE\\
\midrule
DRIFT & 3 & $\best{0.5780\pm0.0003}$ & $\best{10.249\pm0.012}$ & $\best{0.7425\pm0.0007}$ & $\best{10.391\pm0.013}$ \\
Transformer-action & 3 & $\second{0.5790\pm0.0009}$ & $\second{10.281\pm0.002}$ & $\second{0.7432\pm0.0005}$ & $\second{10.421\pm0.005}$ \\
\tftobj{} & 3 & 0.5973 $\pm$ 0.0008 & 10.289 $\pm$ 0.008 & 0.7547 $\pm$ 0.0007 & 10.434 $\pm$ 0.008 \\
TFT-action & 3 & 0.5800 $\pm$ 0.0005 & 10.290 $\pm$ 0.012 & 0.7441 $\pm$ 0.0010 & 10.429 $\pm$ 0.013 \\
\tftlarge{} & 3 & 0.5805 $\pm$ 0.0016 & 10.291 $\pm$ 0.022 & 0.7452 $\pm$ 0.0009 & 10.431 $\pm$ 0.021 \\
GRU-action & 3 & 0.5809 $\pm$ 0.0008 & 10.332 $\pm$ 0.019 & 0.7465 $\pm$ 0.0018 & 10.475 $\pm$ 0.020 \\
Transformer-no-action & 3 & 0.5819 $\pm$ 0.0003 & 10.344 $\pm$ 0.018 & 0.7470 $\pm$ 0.0019 & 10.488 $\pm$ 0.022 \\
TFT-no-dynamic-action & 3 & 0.5814 $\pm$ 0.0004 & 10.344 $\pm$ 0.010 & 0.7472 $\pm$ 0.0008 & 10.486 $\pm$ 0.009 \\
PatchTST-action & 3 & 0.5840 $\pm$ 0.0014 & 10.345 $\pm$ 0.018 & 0.7474 $\pm$ 0.0008 & 10.487 $\pm$ 0.018 \\
Mask-decay GRU-action & 3 & 0.5826 $\pm$ 0.0003 & 10.359 $\pm$ 0.021 & 0.7482 $\pm$ 0.0011 & 10.505 $\pm$ 0.021 \\
GRU-no-action & 3 & 0.5839 $\pm$ 0.0007 & 10.374 $\pm$ 0.026 & 0.7498 $\pm$ 0.0012 & 10.520 $\pm$ 0.028 \\
iTransformer-action & 3 & 0.5864 $\pm$ 0.0012 & 10.393 $\pm$ 0.023 & 0.7535 $\pm$ 0.0014 & 10.538 $\pm$ 0.022 \\
Persistence & 1 & 0.7791 & 12.583 & 1.1077 & 12.791 \\
\bottomrule
\end{tabular}
}
\caption{Complete locked eICU stay-macro ranking, including the broad observed-entry endpoint metric. Neural values are mean $\pm$ seed SD after equal weighting of 8-, 24-, and 48-hour horizons. Boldface and underlining mark the best and second-best values across the complete roster. Persistence is deterministic, and numerical rank does not replace the prespecified paired inference.}
\label{tab:eicu-full-ranking}
\end{table*}

\subsection{eICU Prespecified Inference and Integrated Trajectories}
\label{app:eicu-inference}

\begin{table*}[!t]
\centering
\scriptsize
\setlength{\tabcolsep}{3.0pt}
\resizebox{\textwidth}{!}{%
\begin{tabular}{rrrrrrrrr}
\multicolumn{9}{l}{\textbf{Panel A: horizon-specific terminal estimands}}\\
\toprule
Horizon & All-valid stays & All-valid windows & Obs.-MAP stays & Obs.-MAP stays (\%) & Direct-MAP windows & Direct-MAP windows (\%) & Obs. endpoint stays & Observed 27-var entries\\
\midrule
8 h  & 8,345 & 372,969 & 8,244 & 98.79 & 325,386 & 87.24 & 8,343 & 2,516,966\\
24 h & 8,337 & 239,449 & 7,896 & 94.71 & 204,838 & 85.55 & 8,225 & 1,584,232\\
48 h & 4,486 & 86,171  & 4,265 & 95.07 & 74,131  & 86.03 & 4,438 & 571,780\\
\bottomrule
\end{tabular}
}%
\par\vspace{4pt}
\begin{tabular}{rrrr}
\multicolumn{4}{l}{\textbf{Panel B: selected leads on the fixed trajectory cohort}}\\
\toprule
Lead & Obs.-MAP stays & Direct-MAP windows & Window coverage (\%)\\
\midrule
1  & 4,398 & 79,334 & 92.07\\
8  & 4,410 & 79,421 & 92.17\\
24 & 4,390 & 78,183 & 90.73\\
48 & 4,265 & 74,131 & 86.03\\
\bottomrule
\end{tabular}
\caption{eICU target-observation denominators. In Panel A, observed-MAP stays have at least one directly observed MAP target contributing to the terminal estimand, direct-MAP window percentages use all-valid windows as the denominator, and the observed endpoint column counts stays contributing at least one directly observed target among the 27 variables. Panel B uses the fixed 4,486-stay, 86,171-window trajectory cohort; across all 48 leads, contributing stays range from 4,265 to 4,413 and direct-MAP window coverage ranges from 86.03\% to 92.33\%.}
\label{tab:eicu-observed-denominators}
\label{tab:eicu-trajectory-observed-denominators}
\end{table*}

\begin{table*}[!t]
\centering
\scriptsize
\setlength{\tabcolsep}{2.7pt}
\resizebox{\textwidth}{!}{%
\begin{tabular}{llrrr}
\toprule
Family & Hypothesis & Difference [95\% CI] & $p$ & $q$\\
\midrule
A primary standard MAP & DRIFT vs TFT-action standard MAP & -0.0412 [-0.0514, -0.0307] & $\leq0.001$ & 0.0010 \\
A primary standard MAP & DRIFT vs \tftlarge{} standard MAP & -0.0422 [-0.0594, -0.0239] & $\leq0.001$ & 0.0010 \\
B observed MAP & DRIFT vs TFT-action observed-only MAP & -0.0378 [-0.0483, -0.0275] & $\leq0.001$ & 0.0010 \\
B observed MAP & DRIFT vs \tftlarge{} observed-only MAP & -0.0399 [-0.0553, -0.0249] & $\leq0.001$ & 0.0010 \\
C action dependence & full-stream DRIFT vs TFT-action difference-in-gap & 0.0166 [0.0078, 0.0286] & $\leq0.001$ & 0.0020 \\
C action dependence & future-only DRIFT vs TFT-action difference-in-gap & 0.0152 [0.0058, 0.0282] & $\leq0.001$ & 0.0020 \\
C action dependence & full-stream DRIFT vs \tftlarge{} difference-in-gap & 0.0178 [0.0030, 0.0289] & 0.0150 & 0.0200 \\
C action dependence & future-only DRIFT vs \tftlarge{} difference-in-gap & 0.0181 [0.0015, 0.0307] & 0.0340 & 0.0340 \\
\bottomrule
\end{tabular}
}
\caption{Eight prespecified eICU contrasts. Negative standard/observed MAP differences favor \method{}; positive difference-in-gap values indicate stronger action-conditioned predictive dependence. BH correction is applied separately within the two-, two-, and four-hypothesis families.}
\label{tab:eicu-tests}
\end{table*}

\begin{table*}[!t]
\centering
\scriptsize
\setlength{\tabcolsep}{2.8pt}
\resizebox{\textwidth}{!}{%
\begin{tabular}{llrrr}
\toprule
Analysis & Metric & Difference (DRIFT $-$ Transformer-action) & 95\% CI & $p_{\mathrm{boot}}$\\
\midrule
8/24/48 stay-macro & Endpoint MSE & -0.00103 & [-0.00268, 0.00098] & 0.261\\
8/24/48 stay-macro & MAP MAE & -0.03142 & [-0.04656, -0.01571] & $\leq0.001$\\
8/24/48 stay-macro & Observed endpoint MSE & -0.00063 & [-0.00160, 0.00027] & 0.187\\
8/24/48 stay-macro & Observed MAP MAE & -0.03007 & [-0.04510, -0.01563] & $\leq0.001$\\
Fixed 48-h cohort & Integrated 1--48-h trajectory MSE & -0.00146 & [-0.00455, 0.00141] & 0.297\\
Fixed 48-h cohort & Integrated MAP MAE & -0.04842 & [-0.06911, -0.02946] & $\leq0.001$\\
Fixed 48-h cohort & Integrated observed 1--48-h trajectory MSE & -0.00052 & [-0.00204, 0.00102] & 0.504\\
Fixed 48-h cohort & Integrated observed MAP MAE & -0.04544 & [-0.06562, -0.02604] & $\leq0.001$\\
\bottomrule
\end{tabular}
}%
\caption{Secondary eICU comparison with Transformer-action, the numerical runner-up for standard MAP. Negative values favor \method{}. The analysis uses the frozen paired-bootstrap outputs and was not added to the prespecified A/B/C multiple-testing families.}
\label{tab:eicu-transformer-secondary}
\end{table*}

\begin{table*}[!t]
\centering
\scriptsize
\setlength{\tabcolsep}{3.4pt}
\resizebox{\textwidth}{!}{%
\begin{tabular}{llrrr}
\toprule
Comparison & Metric & Difference [95\% CI] & $p$ & Stays\\
\midrule
DRIFT vs TFT-action & Integrated 1--48-h trajectory MSE & -0.00124 [-0.00161, -0.00089] & $\leq0.001$ & 4,486 \\
DRIFT vs TFT-action & MAP MAE & -0.04420 [-0.05824, -0.02984] & $\leq0.001$ & 4,486 \\
DRIFT vs TFT-action & Integrated observed 1--48-h trajectory MSE & -0.00047 [-0.00127, 0.00030] & 0.243 & 4,486 \\
DRIFT vs TFT-action & Observed MAP MAE & -0.03893 [-0.05428, -0.02554] & $\leq0.001$ & 4,486 \\
DRIFT vs \tftlarge{} & Integrated 1--48-h trajectory MSE & -0.00293 [-0.00661, 0.00030] & 0.074 & 4,486 \\
DRIFT vs \tftlarge{} & MAP MAE & -0.04180 [-0.06157, -0.02179] & $\leq0.001$ & 4,486 \\
DRIFT vs \tftlarge{} & Integrated observed 1--48-h trajectory MSE & -0.00252 [-0.00508, -0.00023] & 0.025 & 4,486 \\
DRIFT vs \tftlarge{} & Observed MAP MAE & -0.03815 [-0.05775, -0.01804] & $\leq0.001$ & 4,486 \\
\bottomrule
\end{tabular}
}
\caption{Paired integrated 1--48-hour results on the fixed 4,486-stay roster. Each stay is summarized first; the three training seeds and 48 lead hours are then equally weighted. Negative values favor \method{}. These trajectory tests are secondary to the prespecified families in Table~\ref{tab:eicu-tests}.}
\label{tab:eicu-trajectory}
\end{table*}

The fixed-roster results preserve the MAP conclusion across the complete lead range. Relative to TFT-action, tensorized trajectory MSE and both MAP summaries favor \method{}, whereas observed-only multivariate trajectory MSE does not differ clearly. Relative to \tftlarge{}, tensorized trajectory MSE is uncertain, observed-only trajectory MSE favors \method{} in the secondary analysis, and both MAP summaries favor \method{}. No equivalence claim is made from intervals that include zero.

\subsection{eICU Action-Replacement Results and Perturbation Audit}
\label{app:eicu-action}

\begin{table*}[!t]
\centering
\small
\resizebox{\textwidth}{!}{%
\begin{tabular}{llrrr}
\toprule
Protocol & Model & MAP gap & 95\% CI & $p$\\
\midrule
Full stream & DRIFT & 0.1278 & [0.1138, 0.1421] & $\leq0.001$ \\
Full stream & TFT-action & 0.1113 & [0.0977, 0.1248] & $\leq0.001$ \\
Full stream & \tftlarge{} & 0.1100 & [0.0939, 0.1267] & $\leq0.001$ \\
Full stream & \tftobj{} & 0.1357 & [0.1173, 0.1549] & $\leq0.001$ \\
Future only & DRIFT & 0.1327 & [0.1170, 0.1476] & $\leq0.001$ \\
Future only & TFT-action & 0.1175 & [0.1028, 0.1320] & $\leq0.001$ \\
Future only & \tftlarge{} & 0.1146 & [0.0993, 0.1310] & $\leq0.001$ \\
Future only & \tftobj{} & 0.1372 & [0.1187, 0.1570] & $\leq0.001$ \\
\bottomrule
\end{tabular}
}
\caption{eICU MAP action-replacement gaps, defined as shifted-action error minus factual-action error. All models use the same five frozen donor mappings, so perturbation prevalence is shared across rows. Positive values indicate conditional predictive dependence, not causal validity.}
\label{tab:eicu-action-gap}
\end{table*}

\begin{table*}[!t]
\centering
\scriptsize
\setlength{\tabcolsep}{3.2pt}
\resizebox{\textwidth}{!}{%
\begin{tabular}{rrrrrrr}
\toprule
Mapping & Eligible stays (\%) & Changed streams (\%) & Changed action entries (\%) &
8-h future changed (\%) & 24-h future changed (\%) & 48-h future changed (\%)\\
\midrule
1 & 100.0 & 16.69 & 4.60 & 11.23 & 17.57 & 26.70 \\
2 & 100.0 & 16.73 & 4.67 & 11.33 & 17.68 & 26.71 \\
3 & 100.0 & 16.62 & 4.58 & 11.16 & 17.50 & 26.52 \\
4 & 100.0 & 16.70 & 4.58 & 11.23 & 17.65 & 26.70 \\
5 & 100.0 & 16.81 & 4.65 & 11.32 & 17.82 & 27.12 \\
\bottomrule
\end{tabular}
}
\caption{eICU donor-mapping audit. All five mappings are distinct, one-to-one, no-self derangements with identical 100\% eligibility under exact time/transition-mask strata. Sparse exposure causes many all-zero-to-all-zero assignments.}
\label{tab:eicu-donor-audit}
\end{table*}

Across the five mappings, 3,234 stays (38.75\%) have a changed complete action stream at least once and 752 (9.01\%) change in every mapping. Each individual mapping changes approximately 1,387--1,403 complete streams. Full-stream and future-only files were required to have protocol order \{\texttt{full\_stream}, \texttt{future\_only}\}, five mappings, horizon order \{8,24,48\}, exact stay identity, and shifted window counts equal to factual counts on every eligible stay; all requirements were satisfied.

\subsection{Cross-Database Replication Implementation and Interpretation Audit}
\label{app:eicu-audit}

The eICU replication used train-only preprocessing and factual-validation checkpoint selection under a fixed evaluation protocol. The sealed analysis evaluated all 36 neural models and Persistence without test refitting. Every factual, trajectory, and action result retains subject, hospital-admission, stay, model, and seed identifiers. Paired statistics require exact roster equality and fail on duplicate keys, nonfinite metrics with positive denominators, denominator mismatches, unexpected action axes, or window-count differences.

The external result supports the following claims: \method{} has the lowest locked-test MAP MAE and the numerically lowest endpoint MSE among the evaluated models; its small MAP advantage versus TFT-action and \tftlarge{} persists on directly observed targets and the fixed 1--48-hour roster; and its action-replacement gaps exceed those of the two direct TFT comparators under both protocols. It does not support claims that \method{} wins every metric, maximizes raw action sensitivity, identifies a treatment effect, validates patient counterfactuals, generalizes to unseen hospitals, or provides a clinically meaningful 0.04-mmHg benefit. In particular, the descriptive audit verifies that all 131 test hospitals and all 212 test hospital--ward pairs were present in training. The snapshot-derived eICU action and the large fraction of unchanged all-zero assignments must remain explicit in any presentation of the external results.

\end{document}